%% file: main.tex
\documentclass{ieeetj}
\usepackage{cite}
\usepackage{amsmath,amssymb,amsfonts}
\usepackage{algorithmic}
\usepackage{graphicx,color}
\usepackage{textcomp}
\usepackage[dvipsnames]{xcolor}
\usepackage{hyperref}
\hypersetup{hidelinks=true}
\usepackage{algorithm,algorithmic}
\def\BibTeX{{\rm B\kern-.05em{\sc i\kern-.025em b}\kern-.08em
    T\kern-.1667em\lower.7ex\hbox{E}\kern-.125emX}}
\AtBeginDocument{\definecolor{tmlcncolor}{cmyk}{0.93,0.59,0.15,0.02}\definecolor{NavyBlue}{RGB}{0,86,125}}

\let\labelindent\relax

\input{_preamble.tex}

\input{_glossary.tex}

\input{_math_notation.tex}

\newcommand{\highlight}[1]{\textbf{\textcolor{NavyBlue}{#1}}}

\def\OJlogo{\vspace{-4pt}$<$Society logo(s) and publication title will appear here.$>$}
\def\seclogo{\vspace{10pt}$<$Society logo(s) and publication title will appear here.$>$}

\def\authorrefmark#1{\ensuremath{^{\textbf{#1}}}}

\begin{document}
\receiveddate{31 October, 2024}
\reviseddate{XX Month, XXXX}
\accepteddate{XX Month, XXXX}
\publisheddate{XX Month, XXXX}
\currentdate{XX Month, XXXX}
\doiinfo{XXXX.2025.1234567}

\markboth{}{Mattamala {et al.}}


\title{Building Forest Inventories with Autonomous Legged Robots --- System, Lessons, and Challenges Ahead}

\author{Mat{\'i}as Mattamala\authorrefmark{1}, 
        Nived Chebrolu\authorrefmark{1},
        Jonas Frey\authorrefmark{2,3},
        Leonard Frei{\ss}muth\authorrefmark{1,4},
        Haedam Oh\authorrefmark{1},
        Benoit Casseau\authorrefmark{1},
        Marco Hutter\authorrefmark{2},
        Maurice Fallon\authorrefmark{1}}

\affil{Oxford Robotics Institute, Department of Engineering Science, University of Oxford, Oxford, UK}
\affil{Department of Mechanical and Process Engineering, ETH Zurich, Zurich, Switzerland}
\affil{Max Planck Institute for Intelligent Systems, T{\"u}bingen, Germany}
\affil{Technical University of Munich, Munich, Germany}
\corresp{Corresponding author: Mat{\'i}as Mattamala (email: matias@robots.ox.ac.uk).}
\authornote{This work is supported in part by the EU Horizon 2020 Project 101070405 (DigiForest) and a Royal Society University Research Fellowship.}

\begin{abstract}
    Legged robots are increasingly being adopted in industries such as oil, gas, mining, nuclear, and agriculture. However, new challenges exist when moving into natural, less-structured environments, such as forestry applications.
    This paper presents a prototype system for autonomous, under-canopy forest inventory with legged platforms.
    Motivated by the robustness and mobility of modern legged robots, we introduce a system architecture which enabled a quadruped platform to autonomously navigate and map forest plots. Our solution involves a complete navigation stack 
    for state estimation, mission planning, and tree detection and trait estimation. We report the performance of the system from trials executed over one and a half years in forests in three European countries. Our results with the ANYmal robot demonstrate that we can survey plots up to \SI{1}{\hectare} plot under \SI{30}{\minute}, while also identifying trees with typical \gls{dbh} accuracy of \SI{2}{\centi\meter}. 
    The findings of this project are presented as five lessons and challenges. Particularly, we discuss the maturity of hardware development, state estimation limitations, open problems in forest navigation, future avenues for robotic forest inventory, and more general challenges to assess autonomous systems.
    By sharing these lessons and challenges, we offer insight and new directions for future research on legged robots, navigation systems, and applications in natural environments.
    Additional videos can be found in \url{https://dynamic.robots.ox.ac.uk/projects/legged-robots}
\end{abstract}



\maketitle

\section{INTRODUCTION}
\IEEEPARstart{S}{ystematically} mapping forests is an important task in modern forestry~\cite{Brack2001} and environmental sciences~\cite{Calders2020}. These maps can be used to understand how a forest contributes to global biomass estimates, the impact of natural and human-induced hazards, and to enable sustainable management within the forestry industry. Obtaining these maps is a laborious process. While remote sensing techniques such as uncrewed platforms or satellites~\cite{Hall2003} have scaled the extent of this task~\cite{Liang2016}, ground measurements are still required under the canopy, to determine fine-level tree traits or calibrate forest models from aerial data~\cite{Duncanson2020}.

Acquiring ground-level measurements is a laborious process. Traditional techniques used in forestry and environmental sciences rely on tape measures, calipers, and handheld laser rangefinders~\cite{Hamilton1988}. Nowadays, technologies such as \gls{tls}~\cite{Dubayah2000} provide high-accuracy, dense reconstructions of forest plots, further enabling the study of biomass estimates and tree structure. However, collecting \gls{tls} data still requires important human efforts in the field, therefore it is constrained to forest plots pre-defined by foresters~\cite{Duncanson2020}.

Robotic solutions have the potential to scale up the mapping process even further. Core technologies such as \gls{slam} have enabled online, large-scale 3D mapping. Handheld SLAM devices have been used to produce ground-level 3D reconstructions of forest plots~\cite{Miettinen2007}, which in forestry is commonly known as \gls{mls}. Further, small-scale drones~\cite{Chen2020, Liu2022} and wheeled platforms~\cite{Pierzchala2018, Tremblay2020} have also been used for under-canopy forest inventory missions.

Quadruped robots can potentially provide significant advantages to aerial and wheeled platforms in this domain. Legs provide increased mobility compared to wheels, and lower soil impact compared to tracked platforms~\cite{Todd1985}. Quadrupeds can also operate for longer missions ($\sim$\SI{90}{\minute}) compared to drones ($<$\SI{30}{\minute}), which are constrained by their battery size~\cite{Liu2022}. Recent applications of legged platforms in crop fields and vineyards~\cite{Quail2023}, open-field scientific sampling applications~\cite{Qian2017}, have demonstrated their potential for outdoor applications, beyond their current use in industrial sites~\cite{Bellicoso2018}. 

\begin{figure*}[t]
  \centering
  \includegraphics[width=\textwidth]{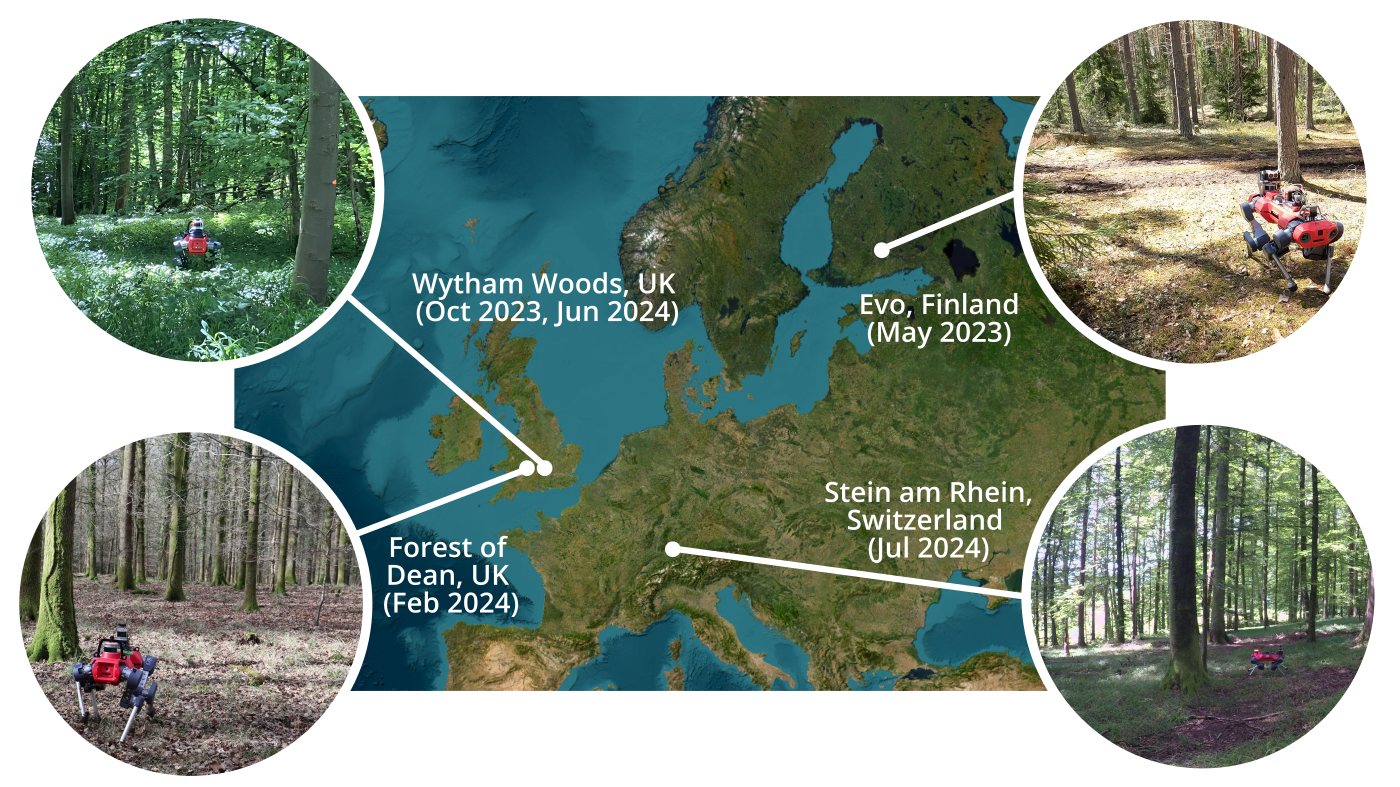}
  \caption{We deployed our autonomous legged robot system in three different countries---Finland, Switzerland, and the United Kingdom---in five campaigns between 2023 and 2024. By testing in different types of forests---mixed, deciduous, and coniferous---across different seasons, we gained valuable insights into the current capabilities of legged platforms for forestry applications.}
  \label{fig:deployments-map}
\end{figure*}

In this work, we study the application of quadruped robots for forestry applications, particularly for \emph{forest inventory}. From a forestry perspective, the ability to autonomously survey forest plots from the ground level can provide data that is not captured from \gls{als} or satellite images. Automating this process with autonomous robots has the potential to facilitate and scale up ground-level surveying, which is otherwise conducted with \gls{tls} and \gls{mls}. While we acknowledge that drones have been the primary platform for these purposes~\cite{Chen2020, Liu2022}, we are interested in understanding how quadrupedal platforms could be used as a complementary platform, testing the maturity of this technology, and identifying core robotics challenges in the forestry context.

Our main contribution is an integrated autonomy system for online forest inventory with quadruped robots and an evaluation of its performance in diverse forest environments. Our solution builds upon a rugged legged platform---the ANYbotics ANYmal, an integrated multi-sensor unit, and state-of-the-art components from locomotion systems to navigation and forest inventory algorithms. We present the development and improvement of the autonomy system, through five field campaigns executed between May 2023 and July 2024 in three European countries---Finland, Switzerland, and the UK---in the context of the DigiForest EU Horizon project\footnote{\url{https://www.digiforest.eu}}, shown in \figref{fig:deployments-map}. On each campaign we deployed a legged platform autonomously in the field, mapping areas up to \SI{1}{\hectare}. 

A publicly available report~\cite{Mattamala2024} presented preliminary results of this study. In this work we extend that analysis with additional results from two new campaigns conducted in Wytham Woods, UK, as well as Stein am Rhein, Switzerland. We report on the the autonomy performance, mapping, and tree inventory capabilities observed in the field. Lastly, we discuss the findings and lessons learned throughout the campaigns.

\section{RELATED WORK}
\label{sec:related-work}
We first review state-of-the-art robotics systems for forestry applications and environmental sciences. Then we focus on the uses of legged platforms for field applications in industrial sites, environmental monitoring, and forestry.

\subsection{Robotics for Forestry and Environmental Applications}
Robotics has the potential to upscale environmental measurement through technologies and platforms that can automate intensive or time-consuming manual labor.

Mapping technologies such as \gls{slam} can accelerate ground-based forest measurements by integrating LiDAR scans from a mobile sensing payload. The produced maps are post-processed to extract the desired attributes, such as the tree positions and \gls{dbh}~\cite{Miettinen2007,Tang2015,Tremblay2020}.
The incremental nature of \gls{slam} and the data acquisition process has motivated the concurrent construction of maps and inventories. Chen~\etal{}~\cite{Chen2020} and Proudman~\etal{}~\cite{Proudman2022}, demonstrated real-time tree detection and \gls{dbh} estimation using small-scale quadcopters and handheld devices, respectively. In this work, we explore this idea by developing an autonomy system that builds upon similar online forest inventory approaches~\cite{Freissmuth2024}.

On the other hand, robotic platforms can support surveying tasks by freeing humans from collecting the data themselves~\cite{Anderson2013a}. In the context of aerial surveying and \gls{als}, autonomous fixed-wing aircraft have been used to obtain LiDAR point clouds \cite{Koh2012}. Other platforms such as small-scale helicopters~\cite{Kellner2019} and more recently hexacopter platforms~\cite{Almeida2020} have been used for over-canopy data collection. Since these platforms fly over the trees, they do not have to deal with challenges such as collision-free navigation. Instead, the questions concern how to survey a defined area, maximizing coverage given energy~\cite{Shah2020} and time constraints~\cite{Schedl2021}.

Using robotic platforms for under-canopy data collection presents more challenges, as it requires robots to operate in cluttered, unstructured spaces. For aerial platforms, it is not possible to use fixed-wing aircraft, hence smaller quadcopters or hexacopters are preferred. Teleoperated small-scale drones have been used for more than ten years for forest surveying~\cite{Lin2011,Hyyppa2020}. More recently, some applications of autonomous drones have been developed for forest inventory~\cite{Chen2020,Liu2022} and DNA sampling~\cite{Aucone2023}. Nevertheless, their small form-factor introduces additional constraints to their capabilities, since the payloads and battery size must be reduced, consequently limiting their autonomy, the complexity of the algorithms, and the extent of the missions they can accomplish~\cite{Zhou2022}. 

An interesting alternative to aerial platforms is tethered systems, such as SlothBot~\cite{Notomista2019}. This has been designed for long-term monitoring of forests with minimal energy requirements. Other approaches combined aerial platforms with a tether to navigate through forest canopies~\cite{Kircheorg2024}, overcoming some of the navigation challenges that a purely aerial platform would face. Nonetheless, many challenges still exist for the implementation and adoption of such systems in monitoring applications.

Ground-based platforms are a competitive approach to providing under-canopy measurements while carrying heavier sensing payloads, enabling longer surveys for forestry or monitoring applications. Wheeled and tracked robot platforms have been used for forest data collection for more than 15 years~\cite{Miettinen2007,Tominaga2018}. These approaches mainly rely on LiDAR \gls{slam} solutions to map the forest~\cite{Pierzchala2018}. Larger-scale wheeled platforms have been demonstrated in the forestry domain to support harvesting operations and achieve autonomous tree cutting~\cite{Rossmann2009,Jelavic2021,Jelavic2022a}. However, ground platforms have larger footprint and mass, and by using rugged wheels and tracks they can damage the soil via trampling effects~\cite{Fountas2010,Calleja-Huerta2023}.

In this work, we propose to use lighter, legged robotic platforms for forestry survey tasks---up to \SI{50}{\kilo\gram}. The focus on smaller platforms is motivated by emerging global forestry policies that advocate for minimal impact, sustainable approaches, such as continuous cover forestry~\cite{EU2021}.

\subsection{Field Applications of Legged Robots}
Since our focus is on legged platforms, we discuss their applications in different fields. From an industrial perspective, legged platforms are being used for inspection and monitoring potentially dangerous or remote sites. For example, they have been deployed in offshore oil and gas facilities~\cite{Bellicoso2018, Ramezani2020offshore, Xin2020}, where their advanced mobility capabilities allow them to navigate complex, human-made environments without requiring the personnel on-site. Nuclear facilities represent another important area where these robots are being deployed~\cite{Staniaszek2024}.

Exploring underground environments is a key application for mining and disaster response. The recent DARPA SubT Challenge~\cite{Chung2023}, held from 2017 to 2021, demonstrated the advantages of legged platforms in overcoming most of the challenges of these unstructured environments. In fact, the majority of the teams participating in the finals used legged platforms as part of their solutions~\cite{Hudson2022, Agha2021}, including the winning team~\cite{Tranzatto2022}.

In open-field applications, legged platforms have been used for agricultural monitoring~\cite{Quail2023, Milburn2023}, litter collection in beaches~\cite{Amatucci2024}, as well as load-carrying tasks~\cite{Wooden2010}. A significant achievement was the autonomous deployment of the LS3 robot over hundreds of kilometers in forests and other natural environments~\cite{Bradley2015}.

For scientific applications, legged platforms have been deployed for soil sampling~\cite{Wilson2021} and measurement of aeolian processes ~\cite{Qian2017}. The K-Rock amphibious robot~\cite{Melo2023} was used for wildlife filming of crocodiles in Africa; the ANYmal robot has also been used to study social interactions with monkeys~\cite{Canteloup2024}. Potential applications for space exploration have also been studied in planetary analog settings~\cite{Arm2023}. 

Forestry applications of legged platforms have been discussed since the 1980s~\cite{Todd1985}. Their potential advantages in terms of heavier payload carrying and lower soil impact have been the main advantages for their use. Despite this potential, the field remains under-explored, with few real-world demonstrations. One example has been in the harvesting process, with the HEAP hybrid wheeled-legged platform~\cite{Jelavic2021}. The other example is more related to our work, involving the deployment of the Boston Dynamics' Spot robot for forest inventory using pre-defined trajectories~\cite{Chirici2023}. In this work, instead, we provide a more extensive experimental demonstration of the capabilities of rugged legged platforms, such as the ANYbotics' ANYmal, for autonomous exploration and inventory in forests.

\section{SYSTEM}
\label{sec:system}

\subsection{Overview}
\begin{figure}[t]
  \centering
  \includegraphics[width=\linewidth]{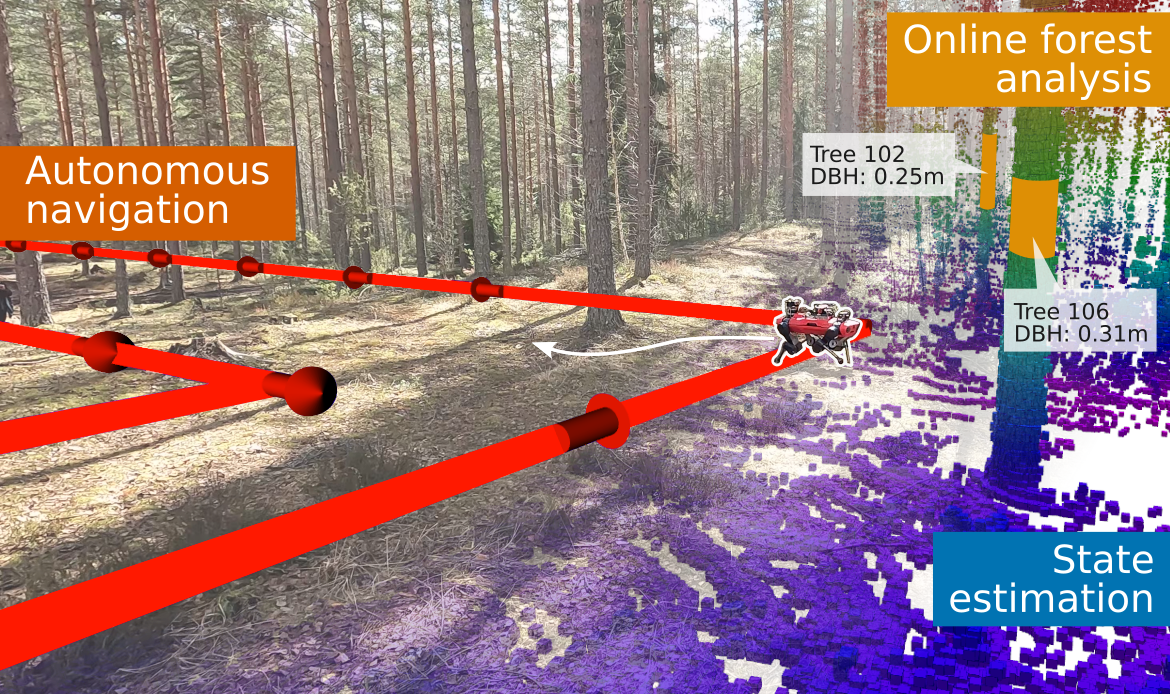}
  \caption{Our vision for autonomous forest inventory with legged robots. Our system autonomously drives a legged platform in an unknown forest plot, while recording environmental data used for tree segmentation and trait estimation.}
  \label{fig:depiction}
\end{figure}

Our goal is to develop a system for autonomous forest inventory using a \SI{50}{\kilo\gram} legged platform. Our vision for this system is presented in \figref{fig:depiction}: A human operator defines an unexplored patch of the forest---a \emph{plot}---they are interested in surveying, which is the input for the mission planning system. Based on this input, the mission planner proposes a preliminary plan for the robot to navigate the environment, given some specifications, e.g., maximizing coverage and consistency (via loop closures).

The robot executes the mission plan using a reactive local planning approach to negotiate obstacles and other unknowns of the environment. While the local planner ensures safe navigation, it also communicates any discrepancies between the planned route and actual surroundings back to the mission planner, enabling dynamic global re-planning of the survey mission as needed. The main data collected by the legged robot consists of 3D scans, which are continuously integrated by the \gls{slam} system running onboard. These scans are used for online forest inventory to determine attributes such as the \gls{dbh} of the surrounding forests, and reported in real-time to the operator, providing live feedback on the mission.

The mission concludes with the robot returning to the starting position. At this point, the complete dataset can then be retrieved, analyzed, and refined in post-processing to generate the forest inventory. The inventory is the main input for the \gls{dss} used by foresters to manage a woodland~\cite{Segura2014}.

\begin{figure*}[b]
  \centering
  \includegraphics[width=0.8\textwidth]{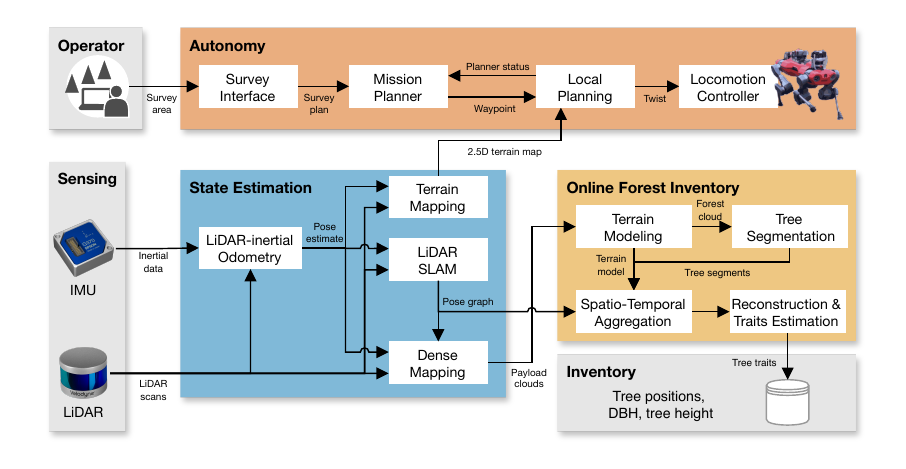}
  \caption{System overview. The \textbf{Autonomy system} executes the mission from a survey reference provided by a human operator. \textbf{State estimation} provides a consistent scene representation and dense clouds used for estimating tree traits. \textbf{Forest inventory} segments the point cloud data and estimates tree traits from point cloud data. The main output of the system is a forest inventory database with the main attributes of the surveyed forest.}
  \label{fig:alf-system-overview}
\end{figure*}

\figref{fig:alf-system-overview} shows the system architecture of our solution, involving state estimation, legged autonomy, and forest inventory. The specific modules are explained in the following sections, and \tabref{tab:mission-summary} provides a summary of their evolution across the campaigns.

\subsection{Robotic Platform}
\label{sec:robotic-platform}
\begin{figure*}[!t]
\centering
  \includegraphics[width=\linewidth]{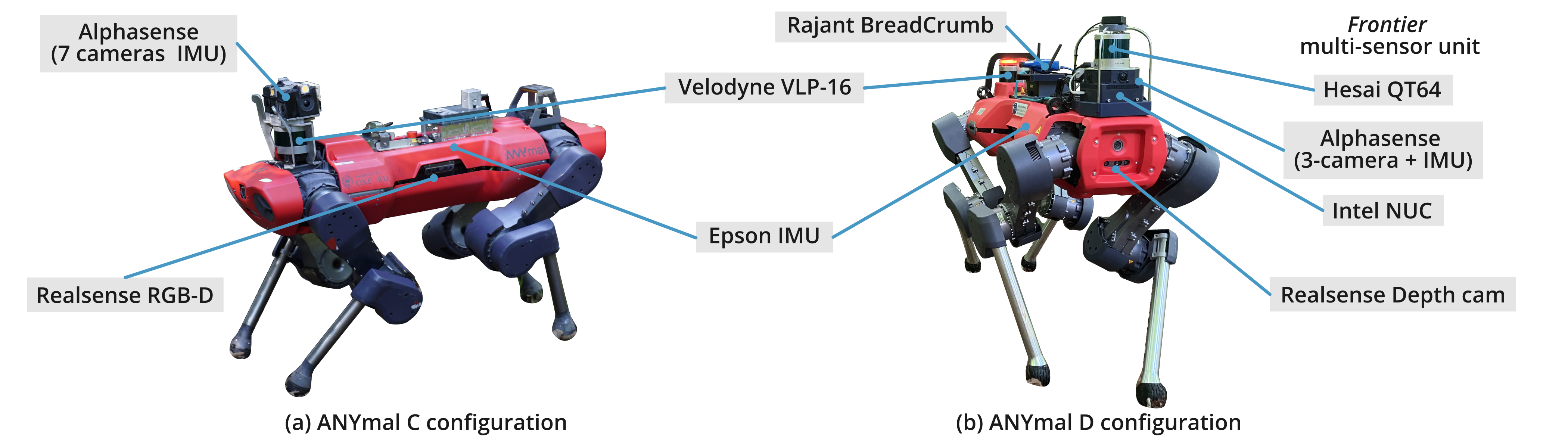}
  \caption{Different configurations of the legged platform used across the campaigns. \emph{Left:} ANYmal C configuration, using the same sensor payload of the DARPA SubT Challenge. \emph{Right:} ANYmal D configuration, using the \emph{Frontier} multi-sensor unit.}
  \label{fig:robot-configurations}
\end{figure*}

In this work, we used the ANYbotics ANYmal platform, specifically \emph{ANYmal C} (released 2020) and \emph{ANYmal D} (released 2022), which are shown in \figref{fig:robot-configurations} along with their main sensors.

For ANYmal C, we used the configuration used by Team CERBERUS in the DARPA SubT Challenge~\cite{Tranzatto2022}. The default sensors (Epson M-G365 \gls{imu}, Velodyne VLP-16 LiDAR, and 4 Intel Realsense cameras) were preserved and an additional Alphasense multi-camera unit was added (7 cameras). Networking was limited to the default WiFi communication antenna. Computing resources on the ANYmal C included two CPUs—--the \gls{lpc} and \gls{npc}---along with a Jetson Xavier GPU. This configuration was used during the 2023 campaigns.

For ANYmal D, we used the default sensors available on the robot, along with the \gls{lpc} and \gls{npc}. No GPU is available in this configuration. We integrated a multi-sensor payload, \emph{Frontier}, which consisted of an Intel NUC CPU as well as a customized Alphasense multi-camera unit (3 cameras) and a Hesai QT64 Wide \gls{fov} LiDAR). Additionally, we introduced a Rajant Breadcrumb module which improved networking in the forest via a mesh network. This configuration has been used for all the campaigns starting in 2024.

Lastly, an additional \gls{opc} laptop was used to run the graphical interface and to command the robot, via WiFi or the Breadcrumb mesh network.

\subsection{State Estimation}
\label{sec:state-estimation}
Our state estimation system has evolved significantly since the first campaign in May 2023. Here, we present our current system used in deployments from late 2023 onward, while the former approach is briefly discussed in the corresponding campaign in \secref{sec:campaigns}.

Our current solution involves four modules: (1) LiDAR-Inertial Odometry, (2) LiDAR SLAM, (3) dense mapping,  and (4) local terrain mapping, which provides representations for forest inventory as well as navigation. Modules (1)--(3) were executed on the \gls{npc} for ANYmal C, or on the Frontier unit for the ANYmal D setup. Module (4) was executed on the Jetson board, or on the \gls{npc}.

\subsubsection{LiDAR-Inertial Odometry}
For odometry, we used a LiDAR-inertial factor graph-based odometry solution, VILENS~\cite{Wisth2023}, which provides high-frequency pose and velocity estimates ($\sim$\SI{200}{\hertz}) in a fixed inertial frame. The system performs sensor fusion of relative pose displacements from LiDAR using the \gls{icp} algorithm~\cite{Besl1992}, aided by preintegrated IMU measurements \cite{Forster2017}.

We did not use proprioceptive sensing from joint encoders in our estimator, as we observed that the LiDAR-inertial approach was effective in forest environments. The tree distribution and the ground points provided sufficient constraints to localize the robot reliably for navigation purposes.

\subsubsection{LiDAR SLAM}
We used a factor graph-based LiDAR SLAM system~\cite{Ramezani2020a} to obtain a consistent estimate of the robot pose in a global fixed frame. Our SLAM system incrementally builds a pose graph, where each node stores a LiDAR scan and 6 DoF pose. The edge factors in the pose graph are formed from incremental odometry estimates from the LiDAR-Inertial system, spaced approximately \SI{1}{\meter} apart.

To mitigate drift as the robot navigates, we incorporated loop closure factors. For this, we detect loop closures by checking the most recent scan against those linked to pose graph nodes within a radius of \SIrange{10}{15}{\meter} from the robot. The candidate loop closures are geometrically verified, by carrying out \gls{icp} registration against a local map built from the LiDAR scans of the neighboring nodes using \texttt{libpointmatcher}~\cite{Pomerleau12comp}. The successful candidates are added as additional factors in the pose graph. 

The pose graph is optimized online in an incremental fashion, using the \gls{isam2} algorithm~\cite{Kaess2012}. Since the LiDAR scans are rigidly attached to the poses, the 3D map is automatically corrected when the poses are optimized. However, we observed that while this representation was sufficient for localization purposes, it was not dense enough for detecting trees. We then introduced an additional dense mapping module to provide more complete representations without impacting the performance of the lightweight pose graph \gls{slam}.

\subsubsection{Dense Mapping}
In order to exploit the continuous data stream from the LiDAR, we implemented a local sub-mapping module. This module accumulates all the sensor-rate LiDAR scans within a fixed-distance window, generating dense scans in the sensor frame, which we refer to as \emph{data payloads}. These data payloads are gravity-aligned and published at a lower frequency ($\sim$\SI{0.1}{\hertz}), accumulating clouds for about every \SI{20}{\meter} traveled. 

The data payloads are the main input for carrying out online tree traits analysis. However, since they only provide a partial view of the trees as the robot traverses the forest, we implemented a spatio-temporal integration approach to leverage the data payloads along with the globally-consistent \gls{slam} path. This enables us to obtain dense clouds from multiple viewpoints to avoid bias in the \gls{dbh} estimation. This is explained in detail in \secref{subsec:forest-inventory}.

\subsubsection{Local Terrain Mapping}
In addition to the large-scale mapping modules, we also perform local terrain mapping and use its output as the main representation for local planning. In our deployments with ANYmal C, we used a GPU-based implementation running on the onboard Jetson computer~\cite{Miki2022b}. With ANYmal D we used the default terrain mapping available in the manufacturer's software stack, which runs on the \gls{npc}. In both cases, we used a \qtyproduct{4 x 4}{\meter} 2.5D grid-based representation with a resolution of \SI{4}{\centi\meter}.

\subsection{Autonomy System}
Our autonomy system is hierarchical, with multiple levels, which span from the user interface for the human operator (top level) to the locomotion algorithms (lower level). 

\subsubsection{Survey Interface} 
\begin{figure*}[t]
\centering
  \includegraphics[width=0.75\linewidth]{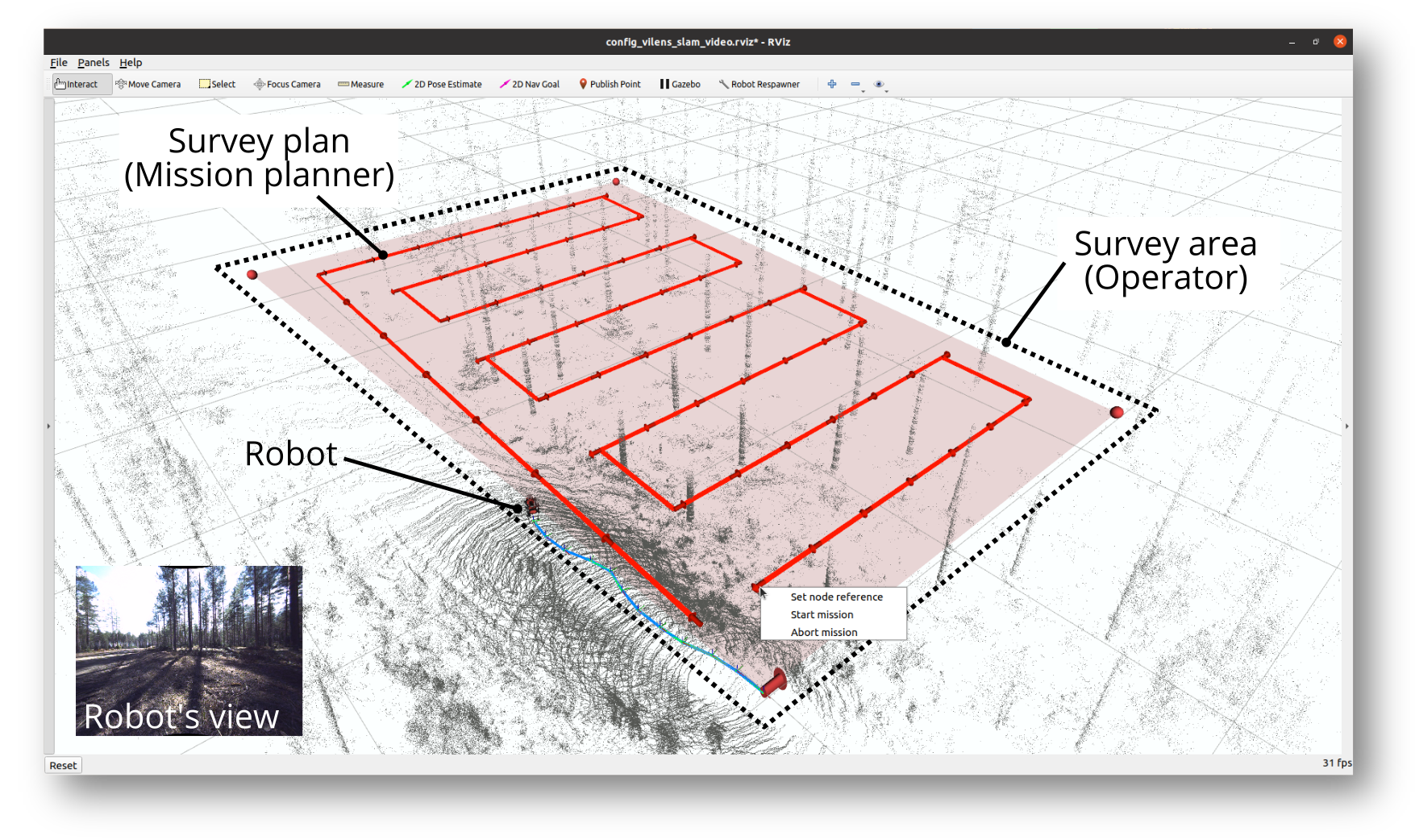}
  \caption{Survey Interface: GUI for the operator, implemented on RViz using interactive markers. The operator defines the area to be surveyed, while the survey plan is automatically proposed by the Mission Planner. The grey point cloud displayed corresponds to the \emph{data payload}.}
  \label{fig:alf-mission-planning}
\end{figure*}

The survey interface operates as the highest level of planning within our system, allowing the human operator to define the robot deployment. We implemented it as a graphical user interface (GUI) on RViz using interactive markers, which runs on the \gls{opc} (\figref{fig:alf-mission-planning}). Given the robot's initial pose, the GUI allows the operator to define a survey area and initialize a \emph{survey plan}, given by a boustrophedon decomposition, commonly known as the lawn mower pattern~\cite{Choset2000}. A similar strategy has been established for \gls{tls} data collection protocols~\cite{Duncanson2020}.

The survey plan is constructed as a sequence of equally-spaced 6 DoF waypoints. The pattern ensures multiple loop closures by enforcing a minimum distance constraint between neighboring path segments, which leads to a more consistent map. 
This plan does not encode the traversability of the plot or the reachability of the waypoints, which is handled by the lower levels of the autonomy system.

The survey interface also provides mechanisms to interrupt the mission at any time, which triggers a safe stopping behavior via the lower levels. It also enables the operator to modify and resume the plan from any specified goal.

\subsubsection{Mission Planner}
The mission planner manages the execution of the survey plan proposed by the operator. It was executed on the \gls{npc} for ANYmal C, and on the Frontier for ANYmal D.

The main task of the mission planner is scheduling the waypoints to be be followed by the local planner, and executing any high-level action from the operator, such as safety stops. We adopt a simple strategy in which the waypoints provide coarse guidance indicating which part of the forest should be explored, and the transitions are executed when the robot is \SIrange{2}{4}{\meter} away from the goal. This allows the local planner to find navigation paths more freely, without requiring a fine alignment to the waypoints' pose.

The mission planner additionally processes signals from the local planner that report the progress towards the goal. In this way, the mission planner can determine if a waypoint is unreachable, and trigger a re-planning strategy to propose alternative waypoints.

\subsubsection{Local Planning}
The local planner generates velocity commands for the low-level locomotion controller to reach a specified waypoint. To achieve this, it relies on the local scene representation built by the local terrain mapping module. 

First, a traversability estimation system processes the local terrain map and extracts geometric features to score the terrain's navigability. The traversability score $s_{\text{trav}}$ is specifically tailored to enable navigation in forest environments with branches and twigs.

This traversability information is used by a reactive local planner~\cite{Mattamala2022} based on \glspl{rmp}~\cite{Ratliff2018}. The approach combines different vector fields, given by analytical expressions or derived from the terrain map. For this application, we compute a continuous cost-to-go $c_{\text{to-go}}$ from the traversability score obtained by the cell-based heuristic:
\begin{equation}
  c_{\text{to-go}} = w_{\text{trav}}\,(1.0 - s_{\text{trav}}) + w_{\text{unkn}}\,s_{\text{unkn}},
\end{equation}
where $s_{\text{unkn}}$ is a fixed score assigned to empty (unknown) cells in the terrain map, and $w_{\text{trav}}$, $w_{\text{unkn}}$ are user-defined weights balancing the influence of known and unknown cells. This cost map is used to compute a vector field for collision avoidance (by thresholding the cost map and computing a \gls{sdf}), as well as a vector field to guide the robot towards the next waypoint (\gls{gdf}).

Simultaneously, the local planner continuously monitors progress toward each waypoint to detect potential unfeasible situations, for example if the proposed waypoint is in the middle of bushes~\cite{Scherer2022}. If a waypoint is determined unreachable, the local planner reports this situation to the mission planner, triggering a re-planning of the next waypoint.

Both the local planner and the traversability analysis modules were always executed on the \gls{npc} of all the versions of the ANYmal we used.
\subsubsection{Locomotion Controller} 
The local planner outputs a 3 DoF velocity command---linear $(x,y)$, angular (yaw)---, which is executed by a low-level \gls{rl}-based locomotion controller. In our experiments, we used both the perceptive controller presented by Miki \etal{}~\cite{Miki2022a}, as well as the blind, learning-based controller from ANYbotics. We used them with no additional modifications to the architecture or training environment to operate in the forest. In both cases, these ran on the \gls{lpc}.

\subsection{Online Forest Inventory}
\label{subsec:forest-inventory}
Our system integrates a state estimation-driven forest inventory pipeline, presented in Frei{\ss}muth \etal{}~\cite{Freissmuth2024}. The approach achieves near real-time tree segmentation and reconstruction during the mission execution. The system was only tested on the ANYmal D campaigns, where it was executed as part of the Frontier software stack.

The main steps are outlined below; for additional technical details, please refer to the original publication~\cite{Freissmuth2024}.

\subsubsection{Terrain Modeling}
The input for the online forest inventory step consists of data payloads generated at \SI{0.1}{\hertz} (see \secref{sec:state-estimation}). For each payload, we extract a terrain model using the \emph{cloth simulation filtering} method by Zhang \etal{}~\cite{Zhang2016b}. Please note that this terrain model is different to the local terrain mapping system used for navigation. Here, the terrain model extracted from the data payload enables us to segment the points belonging to the terrain from the potential tree points.

\subsubsection{Tree Segmentation}
To segment trees, we use a method inspired by Cabo \etal{}~\cite{Cabo2018}. We first normalize the payload cloud z-values using the terrain model, and then filter the data to retain only points within a slice where tree foliage is not expected. We then use a Voronoi-based clustering to obtain tree cluster candidates. We refine the points associated to each tree by fitting a cylinder model and retaining the points within a distance from the cylinder's surface. Finally, the filtered tree points per tree candidates are de-normalized by the terrain model's height to obtain tree candidate clouds in the original reference frame. 

\subsubsection{Spatio-temporal Aggregation}
As mentioned in \secref{sec:state-estimation}, the data payloads provide dense point clouds which are sufficient for initial tree segmentation but they only observe the stems' from a single side. Therefore, the tree candidate clouds cannot be directly used for trait estimation.

To improve the reconstruction, the system aggregates the tree candidate clouds and local terrain models  from different view points over time and space using the \gls{slam} graph. As new tree candidate clouds are determined, we attach them to the closest node in the graph. This enables us to place them in a common coordinate frame, while also correcting their positions when loop closure updates are triggered. By doing this, we can continuously determine when multiple tree candidate clouds correspond to the same tree observed from different viewpoints, or if they correspond to novel instances. Similarly, we update the global terrain model by locally averaging the terrain points along the path. This procedure ensures that even coverage of the trees is achieved before estimating the traits, such as the \gls{dbh}.

\subsubsection{Tree Reconstruction and Trait Estimation}
Lastly, we estimate the traits required by the forest inventory. For this, we \emph{reconstruct} the envelope of the tree trunks, by modeling as as a series of oblique cone frustums. These are obtained by fitting circles along the stem using the spatio-temporally aggregated tree candidate clouds that satisfy certain a minimum angular coverage (we used \SI{90}{\degree}).
The most updated cone frustum models are used to determine the \gls{dbh} and the tree height. The \gls{dbh} is estimated from the cone estimates using the method by Hyyppä~\etal{}~\cite{Hyyppa2020stem}. For the tree height, we use the height of the highest frustum as a coarse estimate. These methods are discussed in detail in \secref{sec:lessons-learned-challenges}.

\begin{figure}[t]
    \centering
    \includegraphics[width=0.9\columnwidth,trim={0 18cm 16cm 7.2cm},clip]{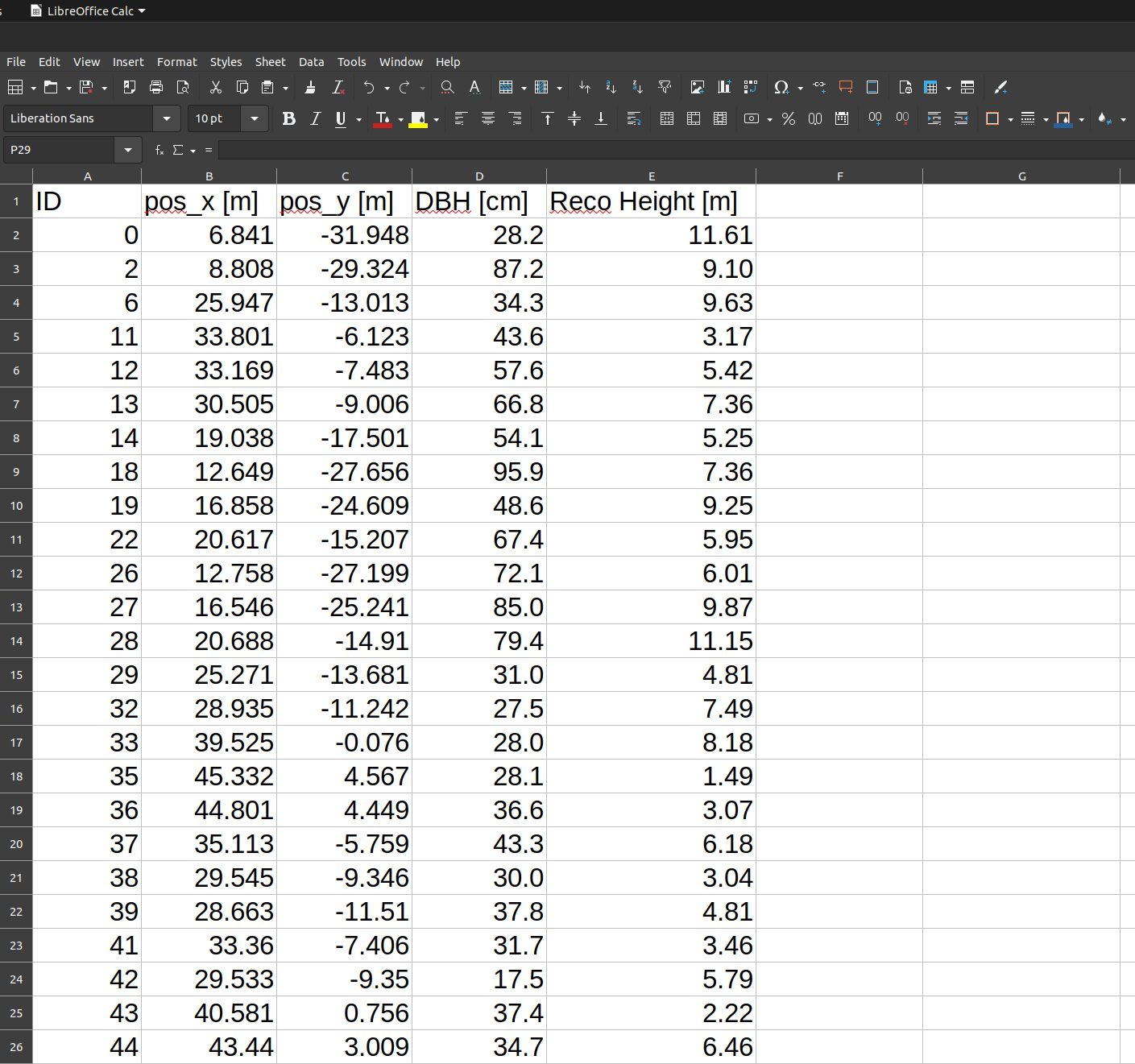}
    \caption{Output forest inventory in spreadsheet format. This example was produced using data from the Forest of Dean campaign. It reports the detected tree positions in a fixed frame, \gls{dbh}, and height of the detected trees.}
    \label{fig:forest-dean-inventory}
\end{figure}

The traits are visually reported to the operator and also exported as a \emph{forest inventory}---a spreadsheet representing the forest as a list of tree locations and associated tree traits (see \figref{fig:forest-dean-inventory}).

\section{FIELD CAMPAIGNS}
\label{sec:campaigns}

\begin{table*}[t]
    \centering
    \footnotesize{
    \begin{threeparttable}
    \input{tables/mission_table}
	\begin{tablenotes}\footnotesize
		\item[$\star$] This mission was manually interrupted.
	\end{tablenotes}
	\end{threeparttable}
    }
    \vspace{10pt}
    \caption{Summary of the campaigns and missions executed between May 2023 and July 2024. We specify the locations, date, and area of the survey defined by the operator. We also report changes in the robot setup (platform, sensing, and software stack). We have \highlight{highlighted} each component that changed with respect to the previous campaign. Refer to \secref{sec:campaigns} for further details.}
    \label{tab:mission-summary}
\end{table*}

\subsection{Campaign description}
Between 2023 and 2024, we executed five field campaigns in different forests in the UK, Finland, and Switzerland. The campaigns consisted of 16 individual missions, which are summarized in \tabref{tab:mission-summary}. 

Five missions (\textsf{Evo-01} to \textsf{Evo-05}) were executed in conifer forests in Evo, Finland in May 2023. Other two were ran in the UK, in mixed (\textsf{WyO-01}, at Wytham Woods, in October 2023) and oak forests (\textsf{Dea-01}, at the Forest of Dean, in February 2024). Three runs during the summer season were executed in a different area of Wytham Woods (\textsf{WyJ-01} to \textsf{WyJ-05}, June 2024), as well as in a mixed forest in Stein am Rheim, Switzerland (July 2024).

\begin{figure*}[t]
    \centering
     \includegraphics[width=\linewidth]{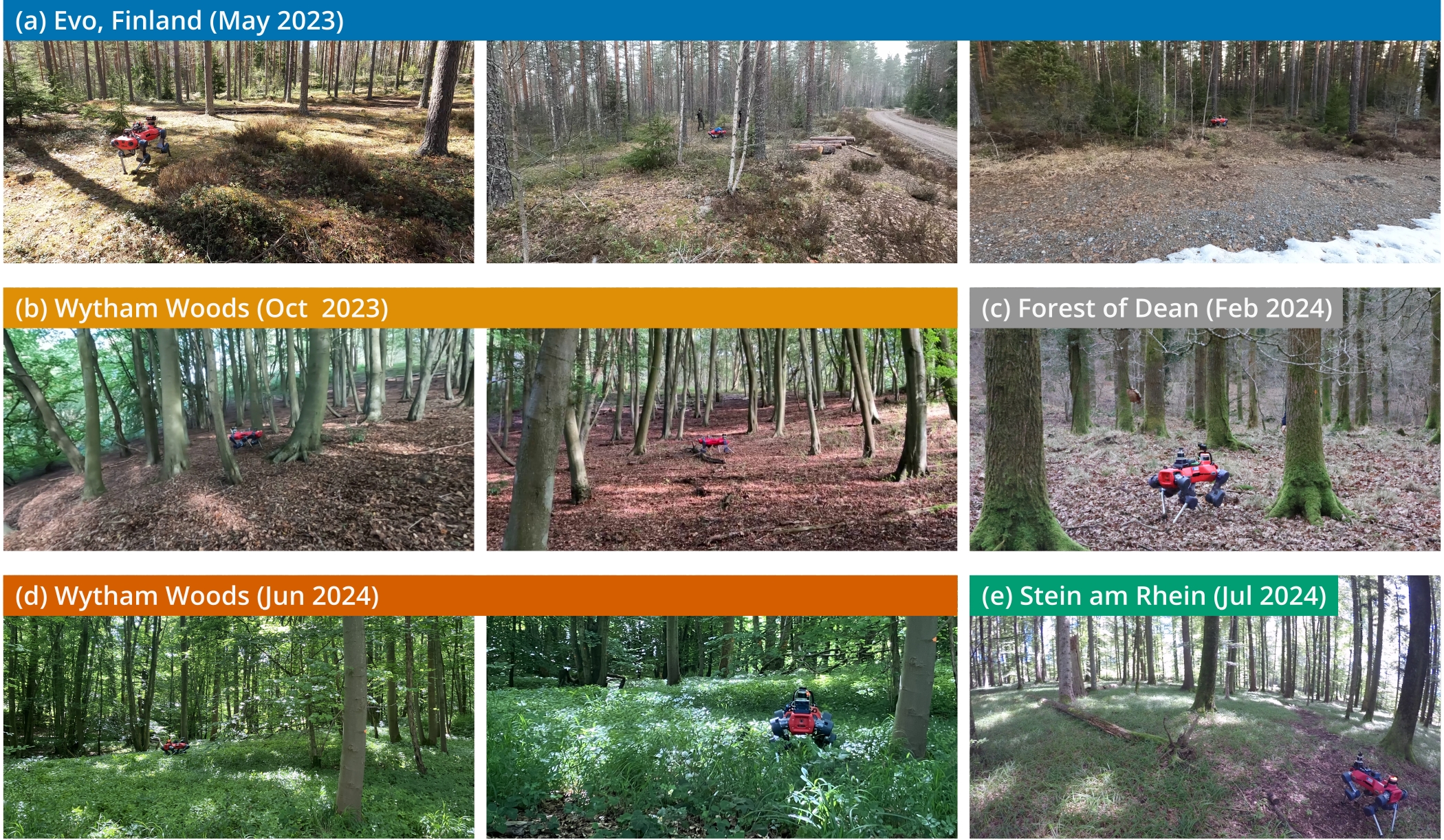}
    \caption{Examples of the autonomous deployments across the campaigns, illustrating the differences in the forest and seasonal conditions. (a) Coniferous forest in Evo, Finland. (b) Mixed forest in Wytham Woods, UK. (c) Oak forest in the Forest of Dean, UK. (d) Mixed forest in Wytham Woods. (e) Mixed forest in Stein am Rhein, Switzerland.
    }
    \label{fig:deployments}
\end{figure*}

The 2023 campaigns developed the autonomy system and in 2024 we focused on the integration of the online forest inventory system. \figref{fig:deployments} illustrates some examples of the robot deployed in the field.

\subsection{Deployment methodology}
All the missions followed the same procedure. A \emph{robot operator} was in charge of defining the survey mission and supervising the overall execution. A \emph{safety operator}, carrying a remote controller to overrule the autonomy system, followed the robot at a safe distance during the mission to intervene and avoid undesired situations (e.g., collisions and \emph{cul-de-sacs}). The interventions were the main signal to evaluate the autonomy system.

\subsection{Campaign 1: Evo, May 2023 (Evo)}
Our first campaign, in Finland, aimed to evaluate the proposed autonomy system in real forest environments. The ANYmal C robot was equipped with the Velodyne VLP-16 (\SI{30}{\degree} \gls{fov}) which is not well suited for forest inventory. We used CompSLAM~\cite{Khattak2020} as the main odometry system. CompSLAM does not perform online pose graph optimization and loop-closure detection to ensure consistency.

Most of the areas in the forest were flat plots with no major undergrowth as the campaign was carried out in late 
autumn (\figref{fig:deployments}a). This simplified the locomotion and navigation challenges as trees and bushes were evident. Missions \textsf{Evo-01} and \textsf{Evo-02} were executed in the same plots, obtaining consistent results. Mission \textsf{Evo-03} was executed in a nearby boggy area. The robot became trapped in a bog until the mission was eventually terminated (\figref{fig:failures}). Missions \textsf{Evo-04} and \textsf{Evo-05} were executed in different, larger areas of the forest. The missions lasted between \SIrange{10}{20}{\minute}, with the robot walking up to \SI{600}{\meter} on the longest missions.

\subsection{Campaign 2: Wytham Woods, October 2023 (WyO)}
The second campaign was executed in a mixed forest plot in Wytham Woods, Oxford, UK. The goal was to improve the state estimation solution problems reported in the Evo campaign. We transitioned to the current state estimation solution of using VILENS for odometry, and VILENS-SLAM for LiDAR SLAM to ensure consistency. The testing area was a sloped plot (\SI{12}{\degree} average inclination) with little undergrowth and a few loose branches, shown in \figref{fig:deployments}b.

\subsection{Campaign 3: Forest of Dean, February 2024 (Dea)}
The third campaign, executed in an oak plantation in the Forest of Dean in the UK, tested the integration of the forest inventory system and the hardware transition to ANYmal~D. We also improved the sensor payload by introducing the Frontier multi-sensor unit~(\secref{sec:robotic-platform}). This switched from the \SI{30}{\degree}  vertical \gls{fov} VLP-16 to the Hesai QT64 LiDAR with a \SI{104}{\degree} \gls{fov}, enabling better coverage of the tree canopy.

This campaign was also executed during winter. Minimal undercanopy and the leaf-fall conditions simplified navigation and reconstruction (\figref{fig:deployments}c). We chose a flat testing area with few boggy patches and no sharp inclines. In this campaign we executed our largest mission, \textsf{Dea-01}, a \qtyproduct{125x30}{\meter} survey area, which corresponded to a \SI{0.93}{\hectare} plot.

\subsection{Campaign 4: Wytham Woods, June 2024 (WyJ)}
Our fourth campaign was carried out during the summer in Wytham Woods. We did not test in the same area as \textsf{WyO-01} because significant undergrowth (about \SI{1}{\meter} high) and dense bushes impeded access. We instead tested in another area, close to an access road, where the undergrowth was about \SI{50}{\centi\meter} tall, see \figref{fig:deployments}d.

For the deployments, we also introduced the Rajant BreadCrumb mesh network. In previous deployments we found the internal WiFi antenna was not sufficient to communicate with the robot and to allow the robot operator to monitor mission execution. With a small network of BreadCrumb units (generally four), we could improve robot-to-base connectivity.

In this campaign, we deployed the robot 5 times (\textsf{WyJ-01}--\textsf{WyJ-05}), aiming to map the same environment every time. For this campaign we changed the locomotion controller to the blind stock controller from ANYbotics, which performed better in the dense undergrowth present in this area.

\subsection{Campaign 5: Stein am Rhein, July 2024 (SaR)}
Our last campaign was executed in Switzerland in the second week of July. The \emph{Oberwald} forest is located beside the northern border with Germany. Much for the forest was untraversable to the robot due to dense foliage---being difficult even for humans to traverse. We selected a plot on a mild slope (\SI{7}{\degree} average) that contained less undergrowth.

The robot was deployed four times (\textsf{SaR-01}--\textsf{SaR-04}) in the chosen forest plot, where it covered only a small region of the area (approximately \SI{0.2}{\hectare}). The objective of these deployments was (1) to evaluate the consistency of the tree detection system, and (2) to demonstrate the system as part of the wider DigiForest research project.

\section{RESULTS}
\label{sec:results}
We report the main results obtained with our system for three different aspects: the autonomy system, the forest inventory performance, and potential future applications. The following sections discuss them in detail, presenting the corresponding evaluation metrics, data, and findings of each case. \tabref{tab:mission-results} provides a summary of the main outcomes.

\begin{table*}[t]
    \centering
    \footnotesize{
    \begin{threeparttable}
    \input{tables/mission_results}
	\begin{tablenotes}\footnotesize
		\item[$\dagger$] This mission was manually interrupted.
            \item[\xmark] Forest inventory system was not available.
	\end{tablenotes}
	\end{threeparttable}
    }
    \vspace{8pt}
    \caption{Summary of the results for all campaigns. We report the autonomy metrics, and the performance of the forest inventory system.}
    \label{tab:mission-results}
\end{table*}

\begin{figure}[t]
    \centering
    \includegraphics[width=\linewidth]{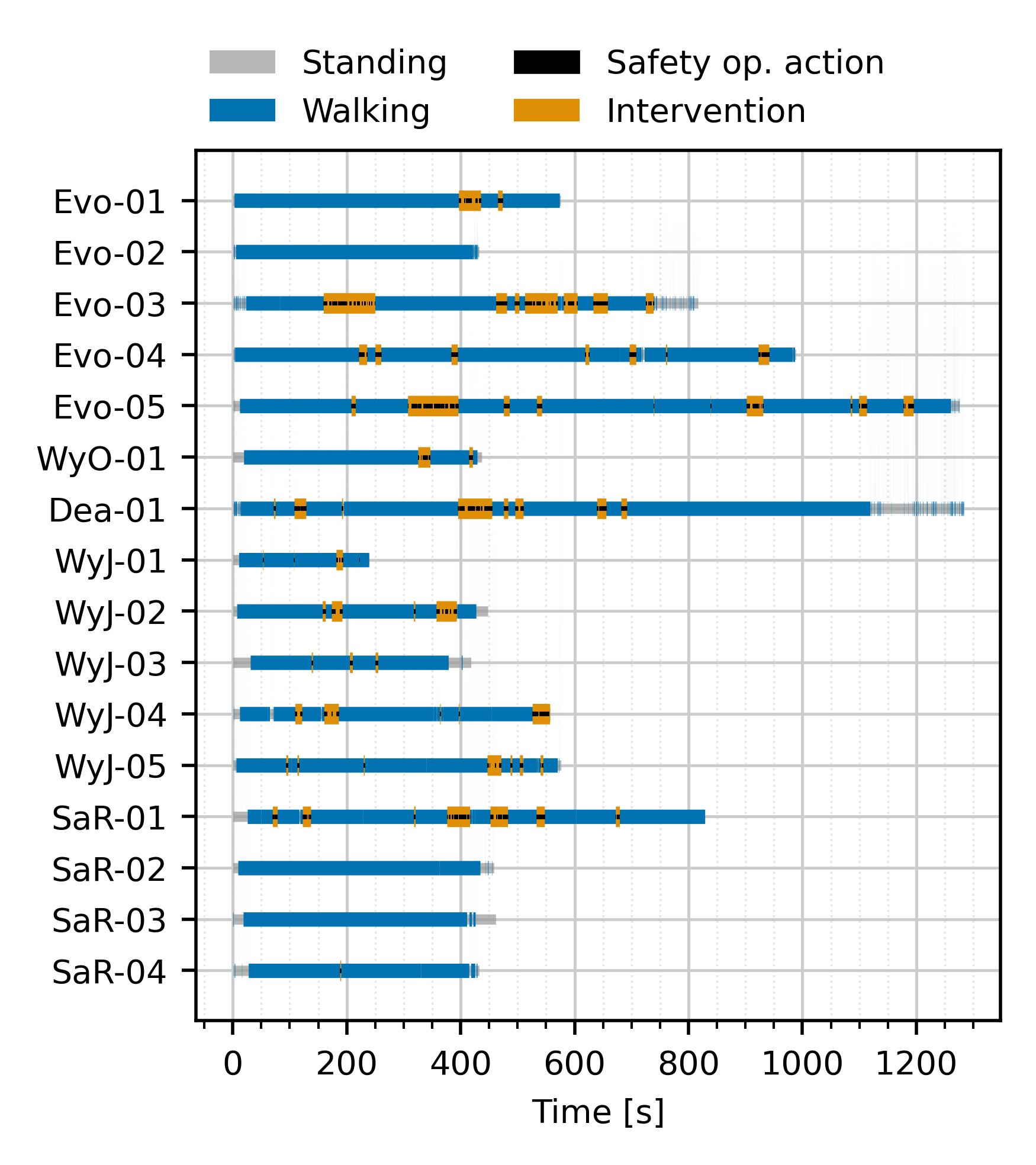}
    \vspace{-10pt}
    \caption{Summary of the autonomy results obtained across the five campaigns. We illustrate the periods where the robots operated fully autonomously (\lightbluesquare{}), and where it was controlled by the safety operator ($\blacksquare$). The aggregated safety actions are defined as \emph{intervention} events (\orangesquare{}).}
    \label{fig:mission-summary}
\end{figure}

\subsection{Analysis of the Autonomy System}
\label{subsec:autonomy-analysis}
Our first analysis focuses on the level of autonomy achieved by our system. To measure this, we used metrics from the autonomous driving literature~\cite{Paz2020}. Particularly, we chose the \gls{mdbi} and \gls{mtbi}, which provide a mechanism to assess the time and distance the robot operated autonomously when compared to the total mission duration and distance traveled.

To compute these metrics, we analyzed the logs of each mission to determine the total time taken on the deployments, as well as the duration of each command given by the safety operator. In principle, each action should be considered an intervention event. However, due to the sparsity of the commands and the variability of the different safety operators to react during the mission, we decided to aggregate all the safety commands within \SI{10}{\second} time windows to define single \emph{intervention} events. This is shown in \figref{fig:mission-summary}.

\begin{figure}[h]
    \centering
    \includegraphics[width=\linewidth]{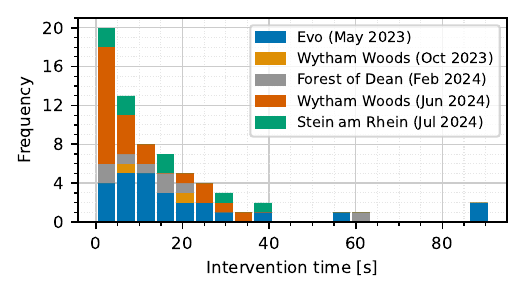}
    \includegraphics[width=\linewidth]{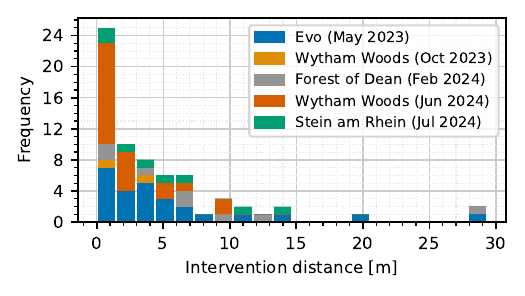}
    \vspace{-20pt}
    \caption{Histogram of the duration (top) and distance traveled (bottom) during the interventions. Most of the interventions reported were short---\SI{20}{\second} on average.}
    \label{fig:hist-time-dist-interventions}
\end{figure}

We matched the intervention signals to the geometrical paths recorded on each mission, which we will denote $\mathcal{M}$ in this section. This allowed us to obtain $N$ time and distance segments in which the robot operated autonomously on each mission, denoted by $d_{\text{auto}}$ and $t_{\text{auto}}$. Then, we computed the \gls{mdbi} and \gls{mtbi} of each mission as follows:
\begin{align}
\text{MDBI}_{\mathcal{M}} & = \frac{1}{N}\sum_{i \in \mathcal{M}}{d_{\text{auto}}^{i}}, \\
\text{MTBI}_{\mathcal{M}} & = \frac{1}{N}\sum_{i \in \mathcal{M}}{t_{\text{auto}}^{i}}.
\end{align}
\tabref{tab:mission-results} reports the number of interventions as well as the \gls{mdbi} and \gls{mtbi} for all the campaigns.
While these metrics have been used to describe the performance of autonomous systems, we observed that the intervention events were usually rare, and short in time. To obtain a better understanding of the interventions events, we additionally aggregated all the intervention events by time and distance to build histograms of their distributions. This is shown in \figref{fig:hist-time-dist-interventions}.

In general, we observed that when all the missions are considered, the interventions follow a Poisson distribution in both time and distance, concentrated around short interventions. The plots indicate that most of the interventions required to manually drive the robot for less than \SI{20}{\second}, or less than \SI{5}{\meter}. A few sparse events in Evo (\textsf{Evo-03}) and the Forest of Dean (\textsf{Dea-01}) required the safety operator to take control for longer periods, specially in the larger scale missions. 

\begin{figure}[h]
    \centering
    \includegraphics[width=\linewidth]{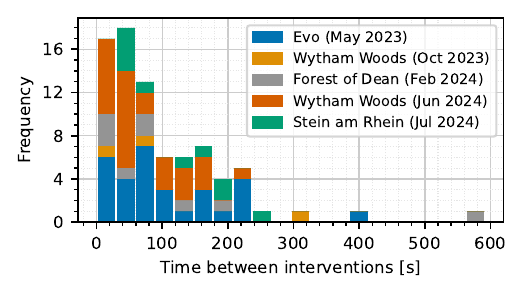}
    \includegraphics[width=\linewidth]{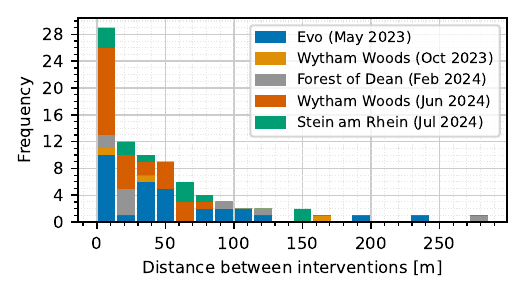}
    \vspace{-20pt}
    \caption{Distribution of time (top) and distance (bottom) between interventions for all missions.}
    \label{fig:hist-time-dist-between-interventions}
\end{figure}

We further revised the \gls{tbi} and \gls{dbi} from a distribution point of view, which is shown in \figref{fig:hist-time-dist-between-interventions}. We removed all the missions where no interventions were reported.

We note that the \gls{tbi} and \gls{dbi} are also Poisson-distributed, reflecting our field observation that the safety operator tended to intervene frequently for short periods of time, only in some parts of the missions. The safety operator's actions mainly aimed to 'help the robot' when the local planner was not able to find a path, or when the mission planner did not produce significant progress by the proposed re-planning strategy. This figure suggests that having a full picture of the distribution provides additional insights, complementary to the summaries provided by the \gls{mtbi} and \gls{mdbi}.

As a general conclusion of this analysis, we acknowledge that since we tested in different conditions with an evolving solution, it is difficult to claim specific achievements in terms of the autonomy of the system. The state estimation stack changed across campaigns, and the differences in the scenery across seasons also introduced new challenges that were not considered in previous missions. 
Nevertheless, the system achieved similar performance across campaigns \emph{on average}. Our solution enabled the robot to navigate autonomously above \SI{80}{\percent} of the distance traveled, or \SI{90}{\percent} of the mission time. The main limitations of our system are discussed in detail in \secref{sec:lessons-learned-challenges}.

\subsection{Analysis of the Forest Inventory System}
\label{subsec:forest-inventory-analysis}
Our second analysis focuses on the performance of the forest inventory system. We report results for the Forest of Dean, Wytham Woods, and Stein am Rhein campaigns.

\tabref{tab:mission-results} summarizes the general metrics of the missions. To estimate the scanning coverage, we considered an \emph{effective range} of \SI{15}{\meter}, which was the maximum LiDAR range to obtain sufficiently dense point clouds for the forest inventory system. Using this value, the estimates of the area covered across the different missions span from \SI{0.16}{\hectare} to \SI{0.93}{\hectare}. The typical scanning speed was \SI{1.8}{\hectare\per\hour}, with the robot walking at an average speed of \SI{0.6}{\meter\per\second} across the different campaigns.

In missions with repeated operation in the same plot, such as Wytham Woods (June) and Stein am Rhein, we detected a consistent number of trees, as reported in \tabref{tab:mission-results}. While we did not have ground truth \gls{tls} measurements for many of the test sites, in our related prior work~\cite{Freissmuth2024} we studied the accuracy of our online inventory system. Therefore, we expect that that the forest inventory can achieve average \gls{dbh} accuracy of \SI{2}{\centi\meter}. This was additionally confirmed in experiments in Stein am Rhein by comparing our estimates against manual measurements with tree calipers. An extract of the forest inventory obtained in mission \textsf{Dea-01} of the Forest of Dean campaign is shown in \figref{fig:forest-dean-inventory}.

\begin{figure*}[t]
    \centering
    \includegraphics[width=\textwidth]{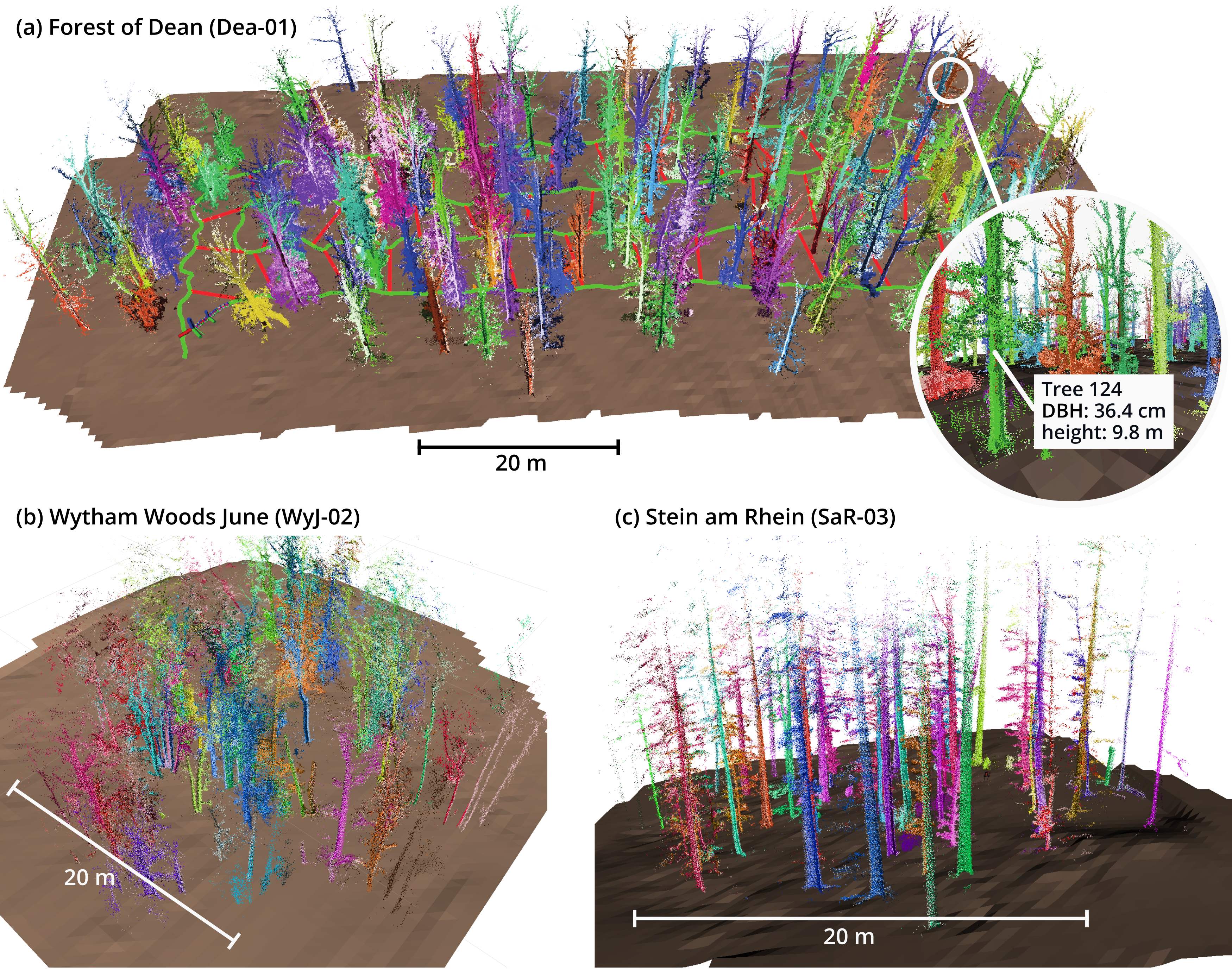}
    \caption{Illustrative example of the different outputs of our forest inventory pipeline in different autonomous deployments. (a) Forest of Dean (\textsf{Dea-01}), (b) Wytham Woods (\textsf{WyJ-02}), (c) Stein am Rhein (\textsf{SaR-03}). We can observe the differences in the tree segmentations due to the diverse species present in the  forests considered.}
    \label{fig:realtime-trees}
\end{figure*}

\figref{fig:realtime-trees} shows visualizations of the forest inventory system when deployed in the Forest of Dean, Wytham Woods (2024), as well as Stein am Rhein. The detected trees are visualized by colored point clouds. The estimated terrain model used to segment the ground is also visualized. We reported differences in the tree species and vegetation level due to the different seasons. In the Forest of Dean, the trees look well-defined, as they were all oak trees without leaves. In Stein am Rhein, it was a mixed forest with a majority of large European beeches, which have a well-defined stem that is easily detected by our system. In contrast, for Wytham Woods tree segmentation was more difficult, mainly due to the mixed species and multi-stem trees that broke some of the assumptions of our tree analysis system.

Our results generally demonstrate the feasibility of achieving online forest inventory with the autonomous legged platform. This is subject to the autonomy performance of the platform, as discussed in \secref{subsec:autonomy-analysis}, as well as the sensing specifications of the \gls{mls} unit being carried. These aspects are further discussed in  \secref{sec:lessons-learned-challenges}.

\subsection{Relocalization in Prior Maps}
\label{subsec:relocalization}

\begin{figure*}[h]
    \centering
    \includegraphics[width=\textwidth]{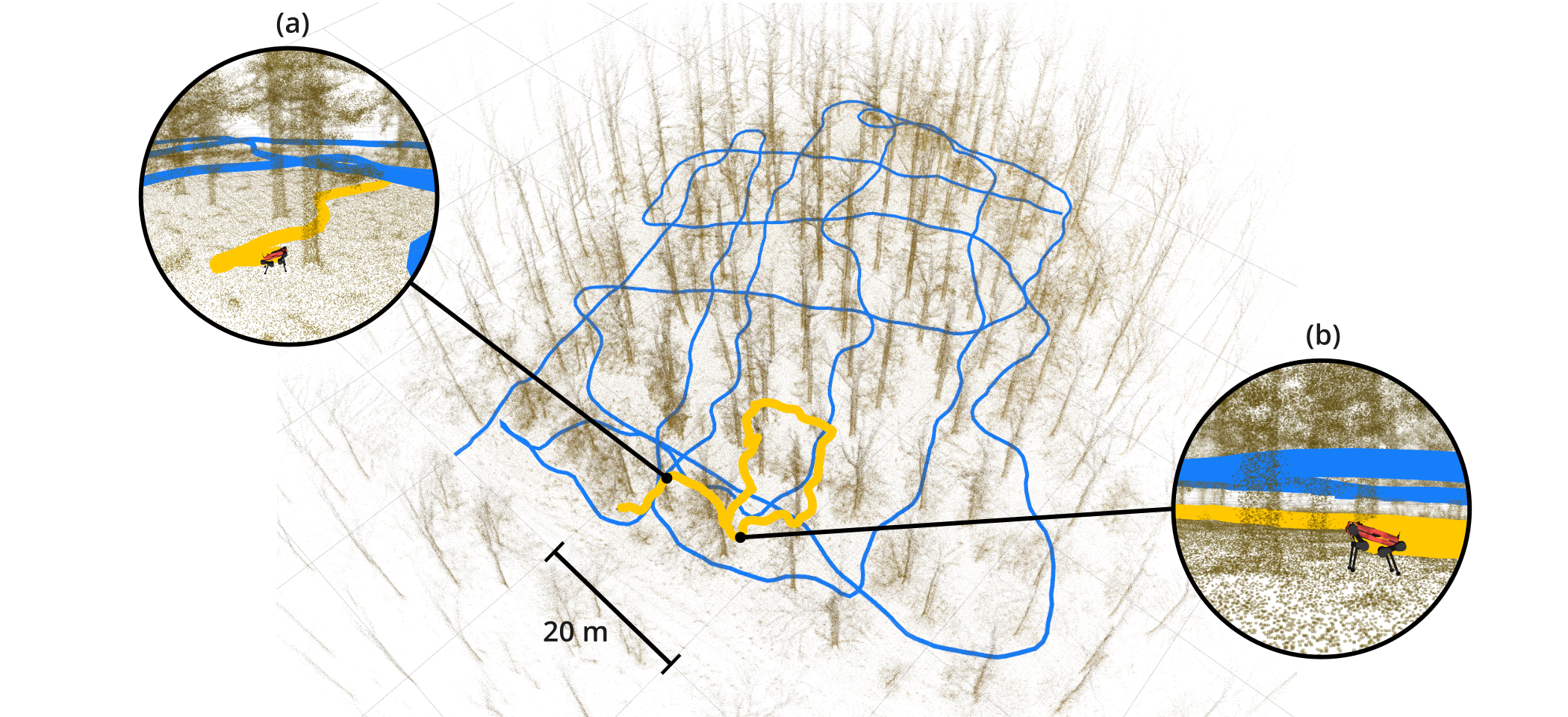}
    \caption{Relocalization example in the Forest of Dean. The blue path \lightbluesquare{} represents the prior trajectory recorded with a \gls{mls} system carried on a backpack at \SI{2.5}{\meter} height. The yellow path \orangesquare{} is executed by the robot in a subsequent localization run. (a) and (b) are close-ups to two segments of the sequence, illustrating the differences between the reference and executed path.}
    \label{fig:forest-dean-relocalization}
\end{figure*}

In our last experiment we report preliminary experiments regarding relocalization in forests with the legged platform for future continuous monitoring. The initial proof of concept of this idea was executed during the Forest of Dean campaign. We collected an initial map using a Frontier unit mounted on a backpack, which was mounted around \SI{2.5}{\meter} above the ground level. A human carried the sensor within the plot to build an initial map, covering an area above \SI{1}{\hectare}.

We additionally recorded three sequences with the ANYmal D platform. We recorded the same signals as done in the other missions reported in the paper, but since the robot is \SI{60}{\centi\meter} height, it had a lower viewpoint with respect to the prior map.

In post-processing, we evaluated a LiDAR-based localization system, which we presented in prior work~\cite{Oh2024}. The system builds upon a forest-specific descriptor for LiDAR place recognition~\cite{Vidanapathirana2022}, which is used to provide global relocalization candidates against the prior pose graph \gls{slam} map. The system only requires as input the LiDAR scans, as well as the LiDAR-inertial odometry estimate. We processed 1 every 4 scans published by the sensor at \SI{10}{\hertz}, effectively producing relocalization estimates at \SI{2.5}{\hertz}.

\figref{fig:forest-dean-relocalization} shows an example of the relocalization sequences. The general overview illustrates the scale of the forest plot. The blue path \lightbluesquare{} indicates the prior trajectory recorded with the \gls{mls} system at \SI{2.5}{\meter} height approximately. The yellow path \orangesquare{} shows the path followed in one of the teleoperated runs, as estimated by our proposed localization system. The differences in the paths' height are further shown in close-ups (a) and (b).

We quantitatively assessed the relocalization system by determining the relocalization rate as well as the distance between relocalizations, which is reported in \tabref{tab:relocalization}. For each of the three sequences, we computed the relocalization rate as the percentage of successfully registered scans with respect to the total number of scans in the sequence. The distance between relocalizations instead was determined by measuring the distance traveled, as estimated by LiDAR-inertial odometry between consecutive relocalizations. We explain the relocalization rate of \SI{20}{\percent} by the strict verification steps we implemented in our method. The distance between relocalizations was below \SI{2}{\meter} (mean and median), with only a few segments where the robot relied on odometry for up to \SI{28}{\meter} before relocalizing again. These preliminary results suggest the potential of our prior work~\cite{Oh2024} to provide reliable localization in monitoring missions.

\begin{table}[t]
    \centering
    \footnotesize{
    \input{tables/relocalization}
    }
    \caption{Summary of the relocalization experiment in the Forest of Dean. We report the number of total scans, the successfully registered ones (\emph{succ.}) and the relocalization rate. Additionally show the summary statistics of the distance between relocalizations, as well as the path length of each sequence (in meters).}
    \label{tab:relocalization}
\end{table}

\section{LESSONS LEARNED AND CHALLENGES AHEAD}
\label{sec:lessons-learned-challenges}
In this section, we discuss the main lessons learned during the five campaigns, as well as the challenges we identified for future research in the field. 

\subsection{Lesson 1 -- Legged Hardware Platform}

Our extensive evaluation supports the conclusions of the DARPA SubT Challenge regarding the hardware maturity achieved by legged platforms in the recent years~\cite{Chung2023}. Current quadruped robot designs have converged to versatile configurations that can deal with many challenges we faced in the field such as branches, twigs, grass, mud, and undergrowth. Additional factors unique to the ANYmal, such as its rugged, IP67 certified design, gave us confidence to deploy it in the wild without risking damage to the robot. 

\textbf{What we learned:} From the perspective of the robotic platform, we demonstrated autonomous coverage up to \SI{1}{\hectare} in \SI{20}{\minute}. This suggests that a robot like ANYmal could reliably carry out inventory missions up to \SI{3}{\hectare} with a single battery charge. This insight helps contextualize how legged robots could compare to other platforms when leveraging battery life and coverage.

From a morphology point of view, we learned that the robot could benefit from longer legs and a different foot design. This was not evident in scenes with scarce undergrowth, but it became more important as the scenes became damp and cluttered, and the undergrowth denser---specially in mixed forests over the summer season. 

\textbf{Challenges ahead:} Our deployments focused on the ANYmal C and D platforms, and prior work also explored the use of the Boston Dynamics Spot robot~\cite{Chirici2023}. Newer commercial platforms such as the Unitree B2 advertise operation up to \SI{5}{\hour}, and research advances in actuators and mechanical design promise even more power-efficient legged robots~\cite{Valsecchi2023, hwangbo2025raibo2}, which can up-scale the extent of the forest inventory missions demonstrated in this work. Furthermore, the ongoing development of new leg designs and feet can also provide morphological advantages for navigating in natural environments~\cite{Godon2024}. 

\subsection{Lesson 2 -- State Estimation and Scene Representations}

In agreement with prior work in the forestry robotics community~\cite{Baril2022, Liu2022}, we showed that state estimations solutions based on LiDAR and inertial sensing provide robust estimates for closed-loop operation of legged platforms in forests. We did not require to introduce additional sensing modalities such as leg odometry or vision to improve the robustness of the state estimation system. However, we note that additional sensing modalities such as vision and haptics could improve the navigation performance, and unlock new traits to determine in the inventory mission.

\textbf{What we learned:} We remark that the success of the LiDAR-inertial setup was not only due to this well-studied sensing setup, but also due to the specific sensors we chose for our sensing payload. Compared to a human or a drone, a legged robot carries the sensors at a lower height (see \figref{fig:forest-dean-relocalization}). We chose a wide \gls{fov} LiDAR mainly motivated by capturing the top of the trees, but also ended up helping the state estimation system work even with tall undergrowth. 
Nevertheless, the payload placement we chose still presents disadvantages for navigation compared to a drone or a human navigating in the forest, where the sensors (or senses) perceive above the undergrowth, providing a better overview of the scene ahead.

\textbf{Challenges ahead:} Building multi-modal scene representations is the main avenue for future research. On the one hand, they can improve the navigation performance by adding scene cues not captured by LiDAR only, such as terrain semantics and traversability information~\cite{Frey2023, Margolis2023}). On the other hand, introducing vision sensing able to robustly operate under canopy, such as \gls{hdr} imaging, can extend the data inputs used in tree trait estimation, opening avenues to detect species and other attributes.

\subsection{Lesson 3 -- Autonomy}

Our exploration strategy for building inventories was motivated by the protocols followed by forest professionals during \gls{tls} and \gls{mls} scanning~\cite{Duncanson2020}. We implemented this approach by separating the problem into different levels, leveraging modules to solve locomotion, local planning, and mission planning. We aimed for a system where the modules interacted with each other to adapt under different situations faced in the field. This was also motivated by prior work on autonomous exploration~\cite{Kottege2023,Tranzatto2023,Scherer2022}.

\begin{figure}[t]
    \centering
    \includegraphics[width=\linewidth]{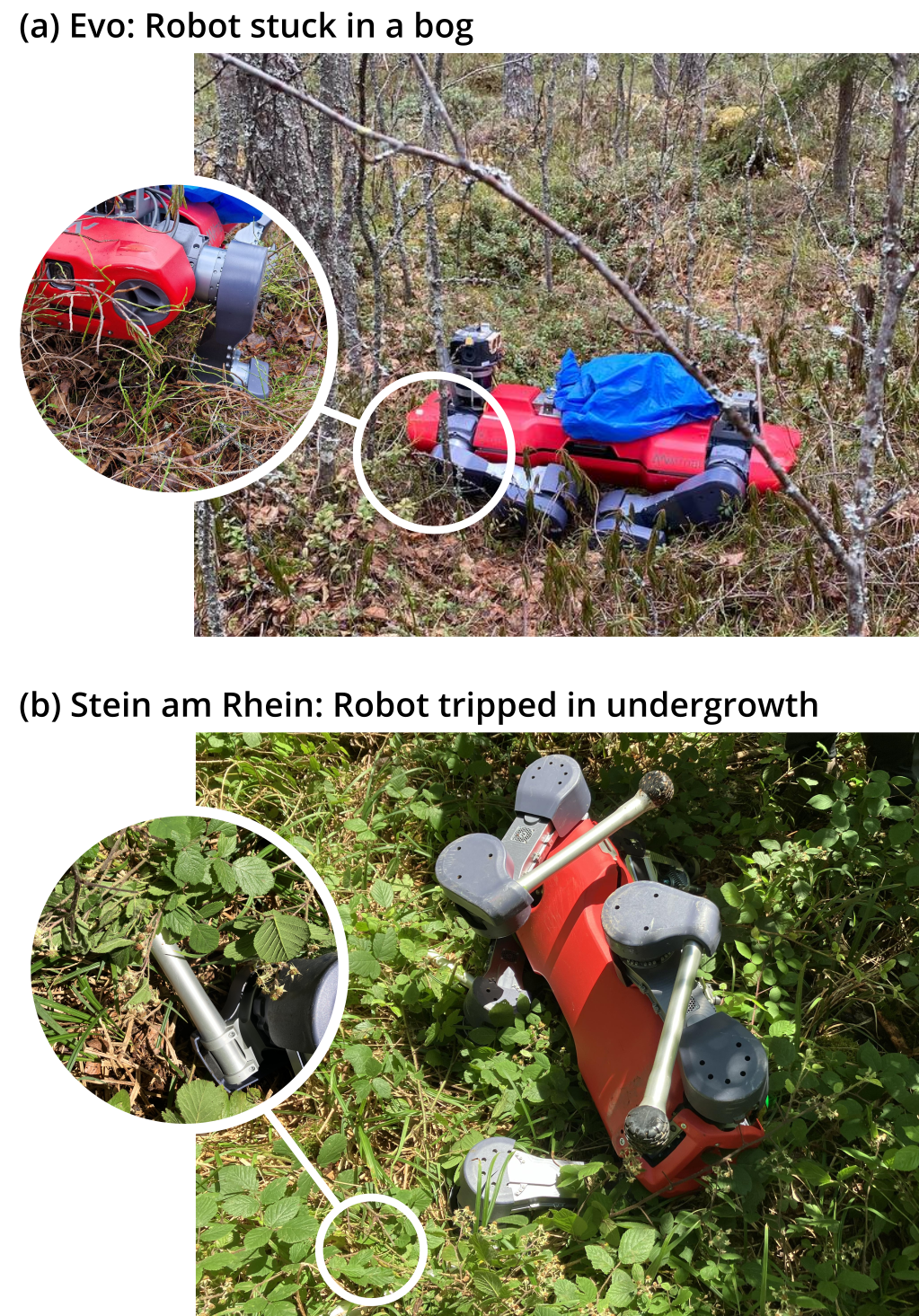}
    \caption{Challenges Ahead: Some examples of failure cases which occurred in Evo and Stein am Rhein. (a) The robot was trapped in bog, were it sank to the body-level and was unable to continue its autonomous mission (\textsf{Evo-03}). (b) The robot fell after tripping due to very dense undergrowth. This occurred while the robot was teleoperated in an area where we were unable to deploy it autonomously.}
    \label{fig:failures}
\end{figure}

\textbf{What we learned:} We point out the local planner as the main module for improvement. The key problems observed were limitations to identify unfeasible waypoints. Part of them were expected due to the limited range of the terrain map, such as goals behind \emph{cul-de-sacs}. However, we also identified situations such as bog-holes or very dense undergrowth areas (see \figref{fig:failures} that would have not been possible to anticipate with LiDAR or vision sensing.

\textbf{Challenges ahead:} Establishing a tighter coupling of the different modules of the autonomy system is the main problem to tackle. By tightening the locomotion and local planning system, it would be possible to navigate more robustly even in situations when not all sensors are suitable, as demonstrated in recent work~\cite{ZhangJin2024}. Adding multi-modal sensing sources is another avenue which could be used to anticipate risky situations and change the mission plan accordingly~\cite{Cai2024}. Furthermore, the mission planning module could be tightly coupled to the forest inventory task, leveraging the tree traits information, energy usage, and time constrain to guide the exploration process, similar to active \gls{slam} approaches~\cite{Placed2023}.

\subsection{Lesson 4 -- Forestry Use-case}

This work aimed to study how legged robots could contribute to forestry inventory tasks, by implementing an autonomous inventory solution following the scanning strategies by forest professionals. Our experiments offer additional evidence on the competitiveness of using \gls{mls} to estimate tree traits, which has already been demonstrated for hand-held scanning~\cite{Chuda2024, Balestra2024} and drone mapping~\cite{Liu2022}. However, the use of legged robots in this problems has not been widely explored~\cite{Chirici2023}, in spite of their potential advantages that have been discussed for as long as 40 years~\cite{Todd1985}.

\textbf{What we learned:} In alignment with the previous lessons, we confirmed that current quadruped platforms could positively contribute to forestry tasks. We foresee them as a complement to current under-canopy drones to map new forest plots, but also as monitoring platforms that could re-visit existing inventories and acquire longitudinal data during day and night.
While we have pointed out technical challenges to be addressed in the previous lessons, the main limiting factor for its adoption might actually be the current cost of these platforms, which is currently comparable to \gls{tls} devices.

\textbf{Challenges ahead:} We acknowledge that the current traits we estimate could be improved (particularly the tree height), and other important traits were actually missing (e.g., species). The former is a challenge that might not be possible to solve only from terrestrial measurements~\cite{Chuda2024}, specially in forests with dense foliage. Combining the output of the forest inventory mission with aerial data can improve the the tree height estimates~\cite{Casseau2024}. Estimating tree species can be addressed by adding additional sensors, such as \gls{hdr} imaging as discussed in Lesson 1.

As also discussed in Lesson 3, a tighter integration of the forest inventory with the autonomy system might also open new exploration strategies that couple the task with the robot capabilities. Similarly, extending the application to other tasks beyond inventory, such as continuous biodiversity monitoring, is another avenue for future research.

\subsection{Lesson 5 -- Assessment}

The nature of the campaigns allowed us to obtain a broad overview of the performance of legged systems across different natural environments. By changing the environment for each trial, we could identify new challenges for the different components of the system. However, we acknowledge that a deeper investigation and assessment of the independent systems should be the next step to overcome the challenges we have raised so far.

\textbf{What we learned:} The main lesson from our diverse deployments is that future developments should mainly target environments with dense undergrowth. These environments raise the most important questions for new perception and mobility research, and use cases where a terrestrial legged robot would be preferable to deploy.

\textbf{Challenges ahead:} How to evaluate complex autonomous systems remains an open question for future research. Bajracharya \etal{}~\cite{Bajracharya2023} set out a comprehensive example of longitudinal assessment. In this work we focused on assessing the autonomous performance, but it would be desirably to explore other metrics to compare aerial platforms, wheeled terrestrial robots, human-carried \gls{mls}, and static \gls{tls} scanning in terms of coverage and scanning time, along with environmental impact (e.g., soil damage) and usability metrics. This is a challenge that requires further work in closer collaboration with foresters and environmental scientists. 

\section{CONCLUSION}
\label{sec:conclusion}

In this work, we presented the development and field deployments of an autonomous forest inventory system using a quadruped robot. Through five campaigns in forests in Finland, the UK, and Switzerland across different seasons, we assessed the feasibility of using a legged robot for this task. The system proposed enabled the ANYmal quadruped to build forest inventories up to \SI{1}{\hectare} in under \SI{30}{\minute}. We analyzed the performance of the system in terms of the autonomy achieved, as well as its potential for forestry and environmental monitoring tasks. 

We summarized the main outcomes of our diverse deployments into five lessons and challenges---discussing hardware, sensing, state estimation, autonomy, and assessment methods. The diversity of our experiments allowed us to provide a general overview of the challenges that quadruped robots might face in natural environments. We recognized local navigation as the main area to improve---which must be aided by robot morphologies, as well as sensing modalities beyond vision and LiDAR. We also argued that future efforts in this domain should focus on natural environments with dense undergrowth, as they currently present the main challenges for autonomous navigation.

The system we have presented is a proof of concept of how legged robots can contribute to forestry tasks, complementing what drones and humans are already able to achieve in this domain. We hope that the insights acquired can motivate future research on legged robots, navigation systems, and applications in natural environments.

\section{ACKNOWLEDGMENTS}
We acknowledge PreFor Oy for organizing the Evo campaign, Forest Research UK for arranging the Forest of Dean campaign, and WSL for their preparations for the Stein am Rhein campaign.
We thank Daniel Marques for his support with the Forest of Dean and Wytham Woods experiments; Wayne Tubby, Matthew Graham, and Tobit Flatscher for hardware integration support; Turcan Tuna for technical support; Ethan Tao, Uljad Berdica, and Frank Fu for field footage; Christina Kassab for feedback on the manuscript.

\bibliographystyle{IEEEtran}
\bibliography{references.bib}


\begin{IEEEbiography}[{\includegraphics[width=1in,height=1.25in,clip,keepaspectratio,trim={0 2cm 0 2cm}]{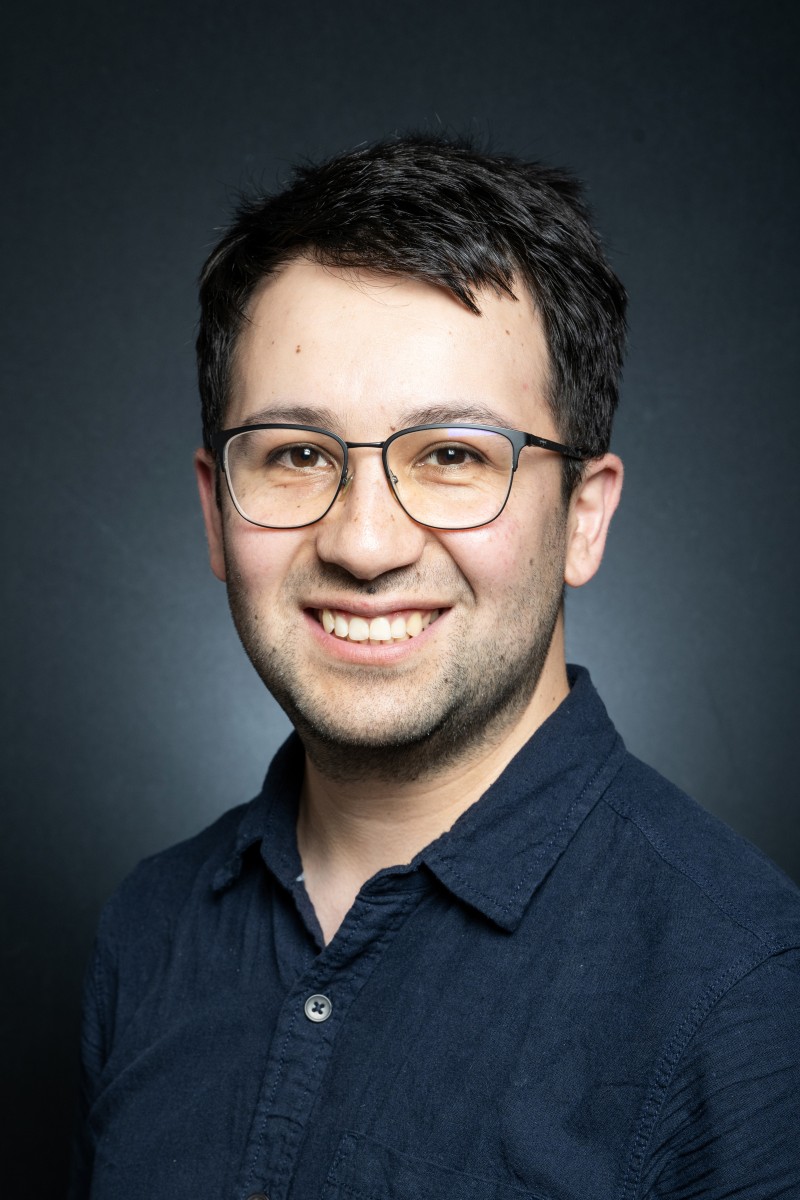}}]
    {Mat{\'i}as Mattamala}~(Member, IEEE)~is a Postdoctoral Researcher in the Dynamic Robot Systems Group at the University of Oxford. He received his M.Sc in Electrical Engineering from the Universidad de Chile in 2018, and his Ph.D. in vision-based legged robot navigation from the University of Oxford in 2023. His research interests are in the foundations and systems for robot autonomy---representations,  perception, action, and learning---with applications to field robotics.
\end{IEEEbiography} 
\begin{IEEEbiography}[{\includegraphics[width=1in,height=1.25in,clip,keepaspectratio]{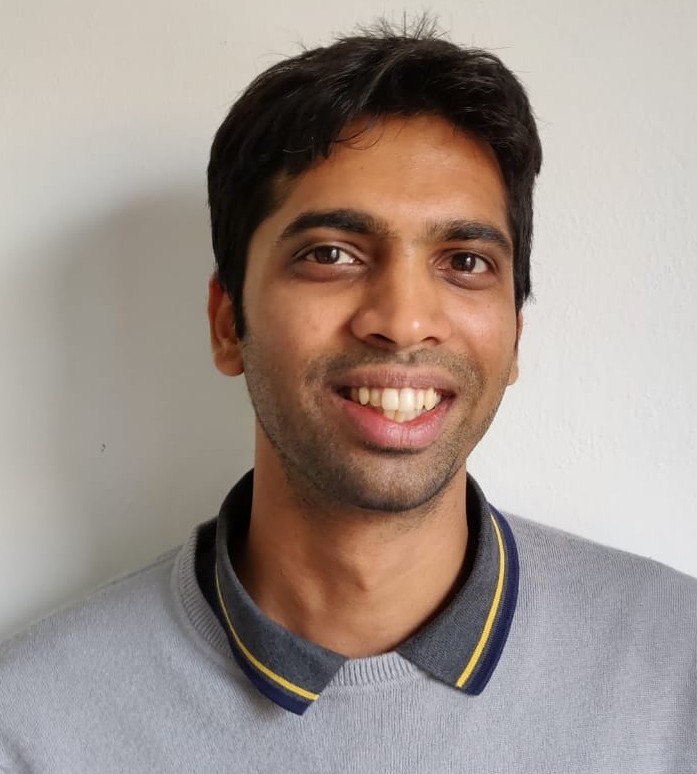}}]
    {Nived Chebrolu}~(Member, IEEE)~is a Postdoctoral Research Associate at the Oxford Robotics Institute, University of Oxford, UK. He received his M.Sc. in Robotics from Ecole Centrale de Nantes (ECN), France, and the University of Genoa, Italy in 2015, and his Ph.D. from the University of Bonn in 2021, where he developed registration techniques for agricultural robotic applications. His research interests are in developing robust localization and mapping techniques for field robotics applications.
\end{IEEEbiography} 
\begin{IEEEbiography}[{\includegraphics[width=1in,height=1.25in,clip,keepaspectratio]{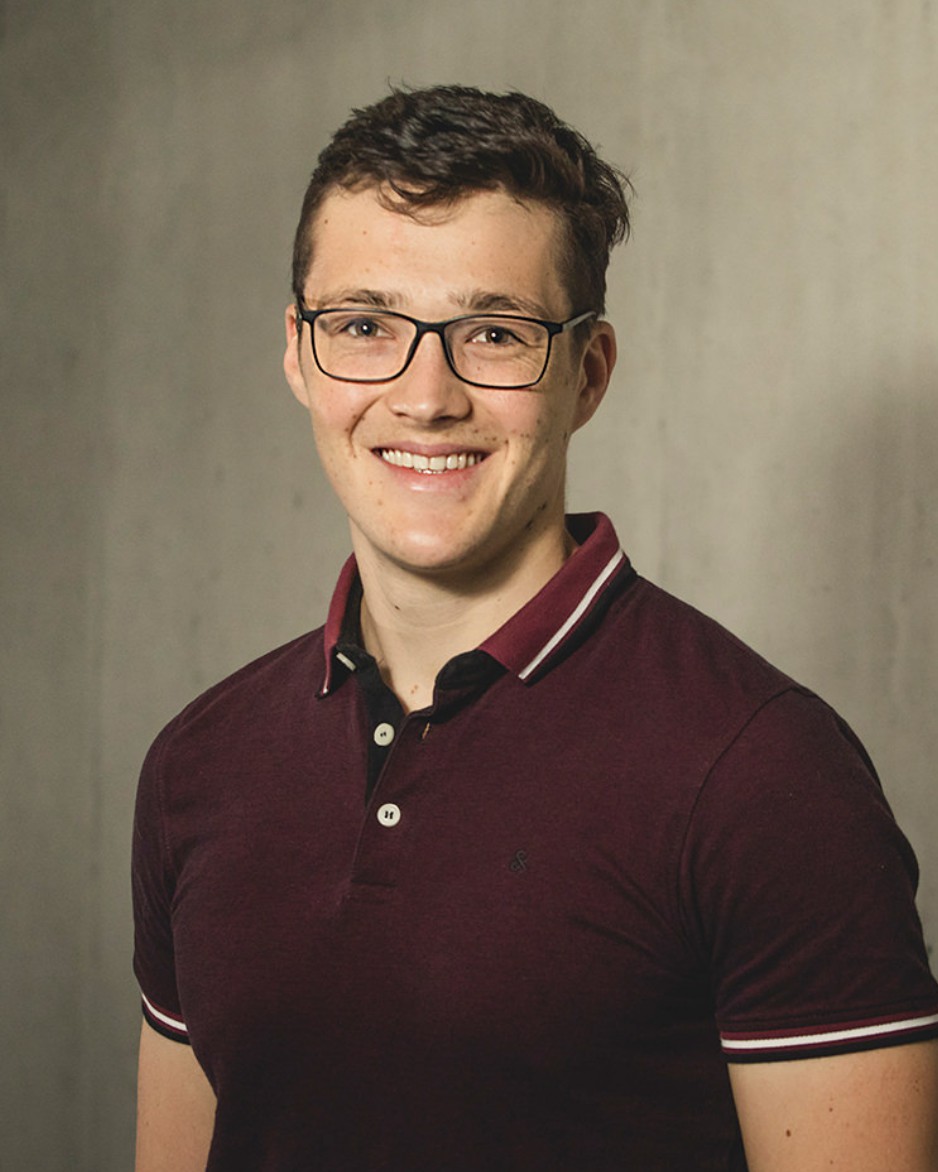}}]
    {Jonas Frey}~(Student Member, IEEE)~is a Ph.D. student in the Robotic Systems Lab at ETH Zurich. He received his M.Sc. in Robotics, Systems \& Control in 2021 from ETH Zurich. He is also affiliated with the Max Planck Institute through the MPI ETH Center for Learning Systems. His research interests lie in the field of perception, navigation and locomotion, and how it can be used for the deployment of mobile robotic systems.
\end{IEEEbiography}
\begin{IEEEbiography}[{\includegraphics[width=1in,height=1.25in,clip,keepaspectratio]{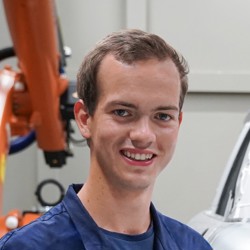}}]
    {Leonard Frei{\ss}muth}~(Student Member, IEEE)~is a is a Ph.D. student in the Smart Robotics Lab at the Technical University Munich, where he also received his M.Sc. in 2024. During his masters thesis he has visited the Oxford Robotics Institute to work on online forest inventory on mobile robots in the context oft the DigiForest EU project. His research interests lie in the field of perception, motion planning, and hardware design, and how the intersection of these fields can yield capable and reliable robots.
\end{IEEEbiography} 
\begin{IEEEbiography}[{\includegraphics[width=1in,height=1.25in,clip,keepaspectratio]{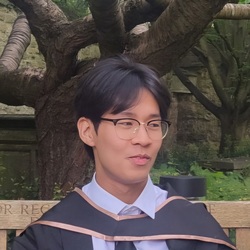}}]
    {Haedam Oh}~(Student Member, IEEE)~is a Ph.D. student in the Dynamic Robot Systems Group at the University of Oxford. He received his MEng from the University of Oxford in 2024. His research interests include localization and mapping with LiDAR and vision.
\end{IEEEbiography}
\begin{IEEEbiography}[{\includegraphics[width=1in,height=1.25in,clip,keepaspectratio]{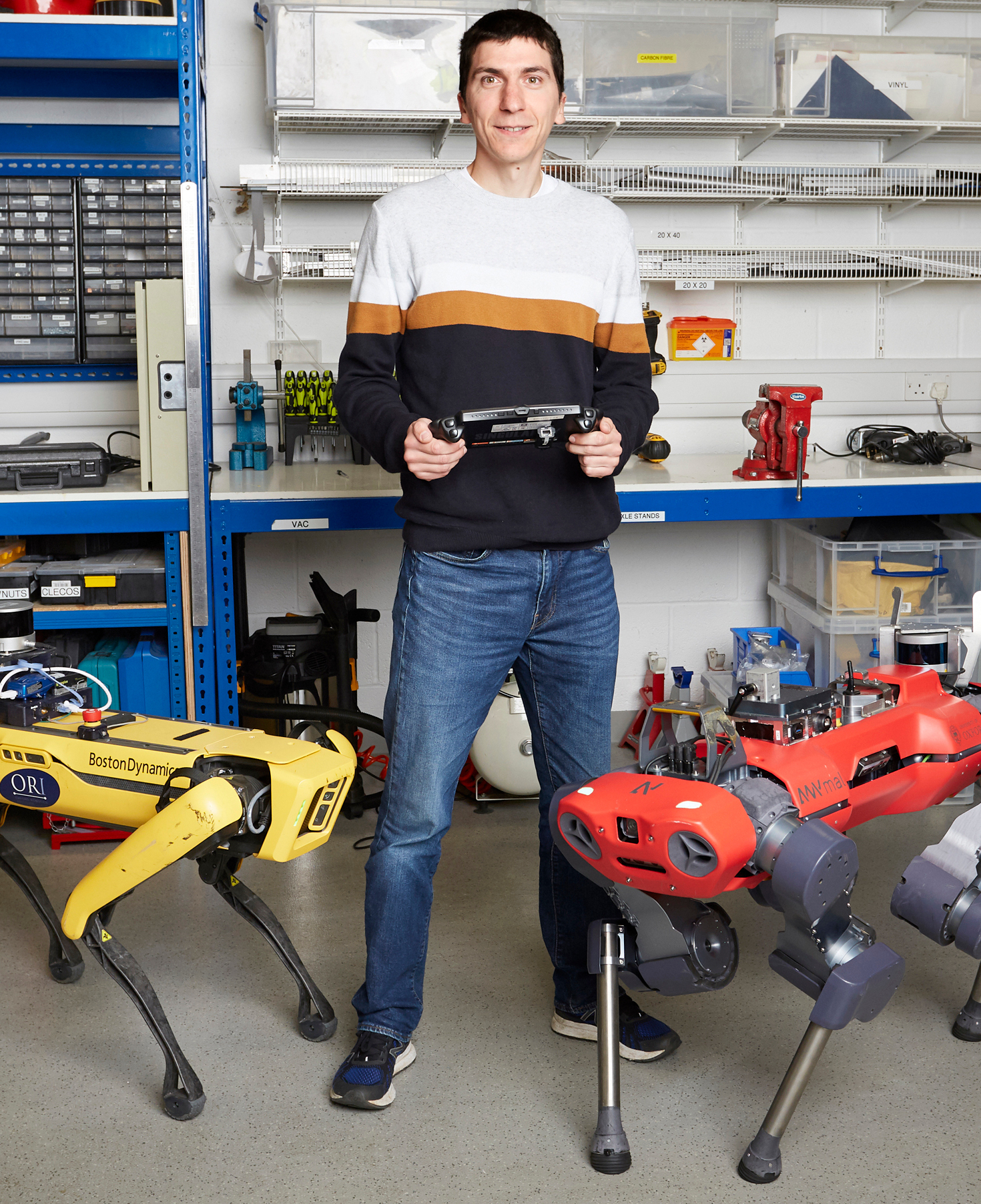}}]
    {Benoit Casseau}~(Member, IEEE)~ is a Robotics Software Engineer at the Oxford Robotics Institute, Oxford, UK. He received his M.Sc. in computer science and applied mathematics in 2013 from {\'E}cole Nationale Sup{\'e}rieure d'Informatique et de Math{\'e}matiques Appliqu{\'e}es, Grenoble, France.  
\end{IEEEbiography} 
\begin{IEEEbiography}[{\includegraphics[width=1in,height=1.25in,clip,keepaspectratio]{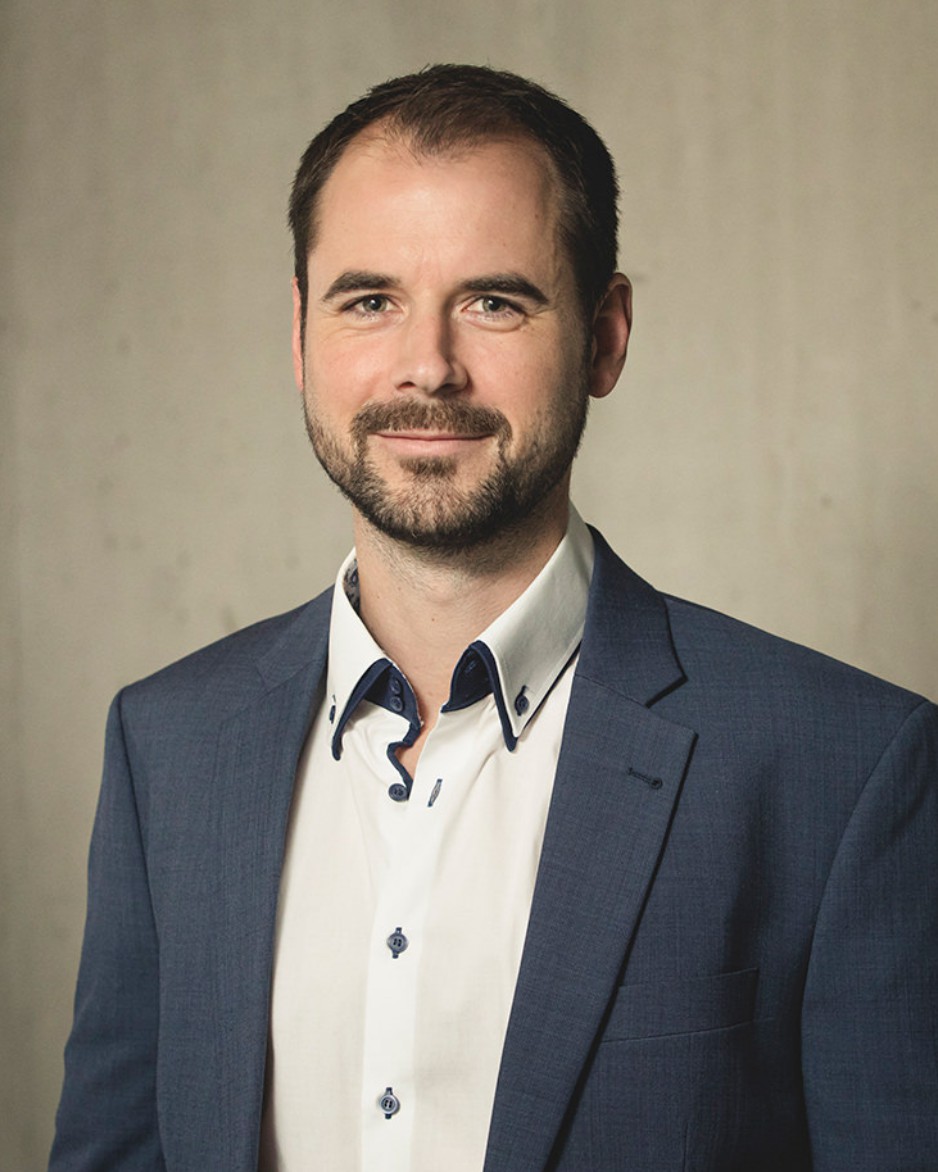}}]
    {Marco Hutter}~(Member, IEEE)~is an Associate Professor for Robotic Systems and Director of the Center for Robotics at ETH Zurich. He received his M.Sc. and Ph.D. from ETH Zurich in 2009 and 2013 in the field of design, actuation, and control of legged robots. His research interests are in the development of novel machines and actuation concepts together with the underlying control, planning, and machine learning algorithms for locomotion and manipulation. Marco is the recipient of an ERC Starting Grant, PI of the NCCRs robotics, automation, and digital fabrication, winner of the DARPA SubT Challenge, and a co-founder of several ETH Startups such as ANYbotics and Gravis Robotics. He is also the Director of the Boston Dynamics AI Institute Zurich office.
\end{IEEEbiography} 
\begin{IEEEbiography}[{\includegraphics[width=1in,height=1.25in,clip,keepaspectratio]{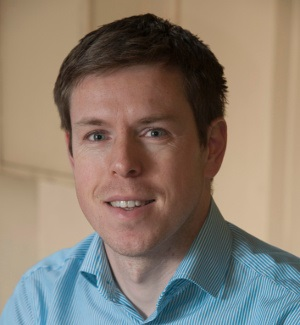}}]
    {Maurice Fallon}~(Senior Member, IEEE)~received the B.Eng. degree in electrical engineering from University College Dublin, Dublin, Ireland, in 2004 and the Ph.D. degree in acoustic source tracking from the University of Cambridge, Cambridge, U.K., in 2008. From 2008 to 2012, he was a Postdoc and a Research Scientist with MIT Marine Robotics Group working on SLAM. Later, he was the Perception Lead of MIT’s team in the DARPA Robotics Challenge. Since 2017, he has been a Royal Society University Research Fellow and an Associate Professor with the University of Oxford, Oxford, U.K. He leads the Dynamic Robot Systems Group, Oxford Robotics Institute. His research interests include probabilistic methods for localization, mapping, multisensor fusion, and robot navigation.
\end{IEEEbiography}

\vfill\pagebreak

\end{document}

%% file: _preamble.tex
\usepackage{times}
\usepackage{amsmath,amsfonts}
\usepackage{algorithmic}
\usepackage{algorithm}
\usepackage{array}
\usepackage{balance}
\PassOptionsToPackage{dvipsnames,table}{xcolor}
\usepackage{textcomp}
\usepackage{stfloats}
\usepackage{url}
\usepackage{verbatim}
\usepackage{graphicx}
\usepackage{todonotes}
\usepackage{booktabs} 
\usepackage{multirow}
\usepackage{multicol}
\usepackage{bbm} 
\usepackage{mathtools}
\usepackage{pgf} 
\usepackage{etoolbox} 
\usepackage{enumitem} 
\usepackage{amssymb}

\usepackage{hyperref} 
\usepackage{lipsum} 
\usepackage[mode=match]{siunitx} 
\usepackage{mathtools}
\usepackage{xspace}
\usepackage[acronym, nonumberlist, nogroupskip]{glossaries-extra}

\usepackage{pbox}

\usepackage{threeparttable}
\usepackage{tablefootnote}

\usepackage{tikz}

\usepackage[normalem]{ulem}
\usepackage{gensymb}

\hypersetup{
colorlinks=true
,linkcolor=NavyBlue
,citecolor=NavyBlue
,urlcolor=NavyBlue
}

\definecolor{high}{RGB}{116, 173, 209}  
\definecolor{low}{RGB}{244, 109, 67}  

\def\secref#1{Sec.~\ref{#1}}

\def\figref#1{Fig.~\ref{#1}}
\def\tabref#1{Tab.~\ref{#1}}
\def\eqref#1{Eq.~(\ref{#1})}

\newcommand{\etal}{et al.}

\usepackage{color}
\definecolor{ColorLightCyan}{rgb}{0.88,1,1}
\definecolor{ColorLightTurquoise}{rgb}{0.5, 1, 0.8}
\definecolor{ColorOrange}{rgb}{1.0, 0.7, 0.0}
\definecolor{ColorTrajBlue}{rgb}{0,0.5,1.0}
\definecolor{ColorTrajOrange}{rgb}{0.87,0.56,0.01}

\usepackage{amssymb}
\usepackage{pifont}
\newcommand{\xmark}{\ding{55}}%

\newcommand{\lightbluesquare}{{\textcolor{ColorTrajBlue}{$\blacksquare$}}}

\newcommand{\orangesquare}{{\textcolor{ColorTrajOrange}{$\blacksquare$}}}

\usepackage{tikz}

\newcolumntype{L}[1]{>{\raggedright\let\newline\\\arraybackslash\hspace{0pt}}m{#1}}
\newcolumntype{C}[1]{>{\centering\let\newline\\\arraybackslash\hspace{0pt}}m{#1}}
\newcolumntype{R}[1]{>{\raggedleft\let\newline\\\arraybackslash\hspace{0pt}}m{#1}}

%% file: _glossary.tex
\robustify{\gls}

\glssetcategoryattribute{acronym}{indexonlyfirst}{true}

\setabbreviationstyle[acronym]{long-short}

\glssetcategoryattribute{acronym}{nohyperfirst}{true}

\newacronym{ape}{APE}{Absolute Pose Error}
\newacronym{apf}{APF}{Artificial Potential Field}
\newacronym{als}{ALS}{Aerial Laser Scanning}

\newacronym{chomp}{CHOMP}{Covariant Hamiltonian Optimization for Motion Planning}
\newacronym{cpm}{CPM}{Camera Performance Model}
\newacronym{cvar}{CVaR}{Conditional Value-at-Risk}

\newacronym{darpa}{DARPA}{Defense Advanced Research Projects Agency}
\newacronym{dbh}{DBH}{Diameter at Breast Height}
\newacronym{dbi}{DBI}{Distance Between Interventions}
\newacronym{dof}{DoF}{Degrees of Freedom}
\newacronym{ddp}{DDP}{Differential Dynamic Programming}
\newacronym{dss}{DSS}{Decision Support System}

\newacronym{fov}{FoV}{Field-of-View}
\newacronym{fmm}{FMM}{Fast Marching Method}

\newacronym{gd}{GD}{Gradient Descent}
\newacronym{gdf}{GDF}{Geodesic Distance Field}
\newacronym{gn}{GN}{Gauss-Newton}
\newacronym{gps}{GPS}{Global Positioning System}
\newacronym{gpmp}{GPMP}{Gaussian Process Motion Planner}
\newacronym{gpu}{GPU}{Graphics Processing Unit}

\newacronym{hdr}{HDR}{High Dynamic Range}

\newacronym{icp}{ICP}{Iterative Closest Point}
\newacronym{imu}{IMU}{Inertial Measurement Unit}
\newacronym{isam2}{iSAM2}{Incremental Smoothing and Mapping}
\newacronym{isp}{ISP}{Image Signal Processor}

\newacronym{komo}{KOMO}{k-Order Motion Optimization}

\newacronym{lagr}{LAGR}{Learning Applied to Ground Vehicles}
\newacronym{lhs}{LHS}{Left-hand Side}
\newacronym{lidar}{LiDAR}{Light Detection and Ranging}
\newacronym{lm}{LM}{Levenberg-Marquardt}
\newacronym{lpc}{LPC}{Locomotion PC}
\newacronym{ls}{LS}{Least Squares}

\newacronym{map}{MAP}{Maximum A Posteriori}
\newacronym{mdbi}{MDBI}{Mean Distance Between Interventions}
\newacronym{mls}{MLS}{Mobile Laser Scanning}
\newacronym{mpc}{MPC}{Model Predictive Control}
\newacronym{mtbi}{MTBI}{Mean Time Between Interventions}

\newacronym{nls}{NLS}{Non-linear Least Squares}
\newacronym{npc}{NPC}{Navigation PC}

\newacronym{opc}{OPC}{Operator PC}

\newacronym{pnp}{PnP}{Perspective-n-Points}
\newacronym{prm}{PRM}{Probabilistic Roadmap}
\newacronym{pte}{PTE}{Pose Tracking Error}

\newacronym{rgb}{RGB}{Red-Green-Blue}
\newacronym[\glsshortpluralkey=RMPs,\glslongpluralkey=Riemannian Motion Policies]{rmp}{RMP}{Riemannian Motion Policy}
\newacronym{rl}{RL}{Reinforcement Learning}
\newacronym{rpe}{RPE}{Relative Pose Error}
\newacronym{rmse}{RMSE}{Root Mean Square Error}
\newacronym{rrt}{RRT}{Rapidly-exploring Random Tree}
\newacronym{ros}{ROS}{Robot Operating System}

\newacronym{sam}{SAM}{Smoothing and Mapping}
\newacronym{sdf}{SDF}{Signed Distance Field}
\newacronym{sfm}{SfM}{Structure-from-Motion}
\newacronym{slam}{SLAM}{Simultaneous Localization And Mapping}
\newacronym{sgd}{SGD}{Stochastic Gradient Descent}

\newacronym{tbi}{TBI}{Time Between Interventions}
\newacronym{to}{TO}{Trajectory Optimization}
\newacronym{tof}{ToF}{Time-of-Flight}
\newacronym{tls}{TLS}{Terrestrial Laser Scanning}
\newacronym{tsif}{TSIF}{Two-State Information Filter}

\newacronym{ugv}{UGV}{Unmanned Ground Vehicle}
\newacronym{urdf}{URDF}{Unified Robot Description Format}

\newacronym{vtr}{VT\&R}{Visual Teach and Repeat}
\newacronym{vit}{ViT}{Visual Transformer}



%% file: _math_notation.tex
\newcommand{\subarrow}[1]{
	\mathord{
		\renewcommand{\arraystretch}{0}
		\begin{array}[t]{@{}c@{}l@{}}
			#1\\[2pt]
			\hspace{-2pt}\scriptstyle\longrightarrow
		\end{array}
		\kern\scriptspace
	}
}


%% file: tables/mission_table.tex
\setlength{\tabcolsep}{3pt}
\renewcommand{\arraystretch}{0.9}
\begin{tabular}{L{1.5cm} L{1.2cm} L{1.5cm} L{1.5cm} L{1.5cm} L{2.5cm} L{2.5cm} L{1.5cm} L{2cm}}
\toprule
\multicolumn{4}{l}{\textbf{Campaign}} & \multicolumn{5}{l}{\textbf{Robot Setup}} \\
\textbf{Location} & \textbf{Mission} & \textbf{Date} & \textbf{Area} [$m^2$] & \textbf{Platform} & \textbf{Sensing} & \textbf{State estimation} & \textbf{Autonomy} & \textbf{Tree Analysis} \\
\midrule
\multirow{5}{1.5cm}{Evo, Finland} & \textsf{Evo-01} & 2023-05-03 & $40 \times 25$ & \multirow{5}{1.5cm}{ANYmal C} & \multirow{5}{2.5cm}{Velodyne VLP-16, Epson IMU} & \multirow{5}{2.5cm}{CompSLAM odometry} & \multirow{5}{1.5cm}{Perceptive locomotion} & \multirow{5}{1.5cm}{None} \\
 & \textsf{Evo-02} & 2023-05-03 & $40 \times 25$ &  &  &  &  &  \\
 & \textsf{Evo-03} & 2023-05-04 & $40 \times 35$ &  &  &  &  &  \\
 & \textsf{Evo-04} & 2023-05-04 & $35 \times 35$ &  &  &  &  &  \\
 & \textsf{Evo-05} & 2023-05-05 & $70 \times 25$ &  &  &  &  &  \\
 \midrule
Wytham Woods, UK & \textsf{WyO-01} & 2023-10-06 & $20 \times 20$ & ANYmal C & Velodyne VLP-16, Epson IMU &  \highlight{VILENS odometry, VILENS-SLAM} & Perceptive locomotion & None \\
\midrule
Forest of Dean, UK & \textsf{Dea-01} & 2024-02-19 & $125 \times 30$ & \highlight{ANYmal D} & \highlight{Frontier (Hesai QT-64)} & VILENS odometry, VILENS-SLAM &  Perceptive locomotion & \highlight{Integrated} \\
\midrule
\multirow{5}{1.5cm}{Wytham Woods, UK} & \textsf{WyJ-01} & 2024-06-25 & $12 \times 12$ & \multirow{5}{1.5cm}{ANYmal D} & \multirow{5}{2.5cm}{Frontier (Hesai QT-64), \highlight{Rajant mesh network}} & \multirow{5}{2.5cm}{VILENS odometry, VILENS-SLAM} & \multirow{5}{1.5cm}{\highlight{Blind locomotion}} & \multirow{5}{1.5cm}{Integrated} \\
 & \textsf{WyJ-02} & 2024-06-25 & $20 \times 15$ &  &  &  &  &  \\
 & \textsf{WyJ-03} & 2024-06-25 & $20 \times 15$ &  &  &  &  &  \\
 & \textsf{WyJ-04} & 2024-06-25 & $22 \times 20$ &  &  &  &  &  \\
 & \textsf{WyJ-05} & 2024-06-25 & $22 \times 15$ &  &  &  &  &  \\
\midrule
\multirow{4}{1.5cm}{Stein am Rhein, Switzerland} & \textsf{SaR-01} & 2024-07-08 & $26 \times 22$ & \multirow{4}{1.5cm}{ANYmal D} & \multirow{4}{2.5cm}{Frontier (Hesai QT-64), Rajant mesh network} & \multirow{4}{2.5cm}{VILENS odometry, VILENS-SLAM} & \multirow{4}{1.5cm}{Perceptive locomotion} & \multirow{4}{1.5cm}{Integrated} \\
 & \textsf{SaR-02} & 2024-07-09 & $16 \times 20$ &  &  &  &  &  \\
 & \textsf{SaR-03} & 2024-07-09 & $16 \times 20$ &  &  &  &  &  \\
 & \textsf{SaR-04} & 2024-07-09 & $16 \times 20$ &  &  &  &  &  \\
\bottomrule 
\end{tabular}

%% file: tables/mission_results.tex
\setlength{\tabcolsep}{3pt}
\renewcommand{\arraystretch}{0.9}
\begin{tabular}{L{1.5cm} C{1.2cm} C{2.2cm} C{1.5cm} C{1.5cm} C{1.5cm} C{2.3cm} C{1.3cm}}
\toprule
\textbf{Mission} & \multicolumn{5}{l}{\textbf{Autonomy}} & \multicolumn{2}{l}{\textbf{Forest Inventory}}\\
\textbf{Name} & \textbf{Time}~[$s$] & \textbf{Dist. traveled}~[$m$] & \textbf{Interv.}~[$\#$] & \textbf{MDBI}~[$m$] & \textbf{MTBI}~[$s$] & \textbf{Area covered}~[ha] & \textbf{Trees}~[$\#$]\\
\midrule
\textsf{Evo-01} & 575.6  & 270.3 & 2  & 84.8  & 176.2 & 0.33 & \xmark \\
\textsf{Evo-02} & 432.0  & 233.6 & 0  & 233.6 & 432.4 & 0.31 & \xmark \\
\textsf{Evo-03}$^\dagger$ & 816.8  & 301.1 & 7  & 30.5  & 72.2  & 0.36 & \xmark \\
\textsf{Evo-04} & 988.4  & 336.6 & 7  & 39.4  & 114.2 & 0.36 & \xmark \\
\textsf{Evo-05} & 1275.5 & 609.7 & 10 & 51.1  & 99.6  & 0.58 & \xmark \\
\midrule
\textsf{WyO-01} & 436.9 & 215.0 & 2 & 69.9 & 136.4 & 0.29 & \xmark \\
\midrule
\textsf{Dea-01} & 1283.5 & 665.4 & 8 & 66.0 & 127.8 & 0.93 & 97 \\
\midrule
\textsf{WyJ-01} & 239.5 & 82.4  & 4 & 15.8 & 45.2  & 0.16 & 28 \\
\textsf{WyJ-02} & 447.7 & 163.3 & 4 & 28.8 & 77.0  & 0.20 & 46 \\
\textsf{WyJ-03} & 418.1 & 139.4 & 3 & 33.9 & 101.6 & 0.20 & 41 \\
\textsf{WyJ-04} & 556.6 & 179.4 & 5 & 27.3 & 80.8  & 0.22 & 46 \\
\textsf{WyJ-05} & 576.1 & 196.9 & 7 & 22.9 & 66.2  & 0.22 & 52 \\
\midrule
\textsf{SaR-01} & 828.7 & 286.5 & 7 & 30.3  & 88.8  & 0.26 & 66 \\
\textsf{SaR-02} & 458.3 & 149.1 & 0 & 149.1 & 458.3 & 0.20 & 43 \\
\textsf{SaR-03} & 461.5 & 152.1 & 0 & 152.1 & 461.5 & 0.20 & 44 \\
\textsf{SaR-04} & 432.4 & 140.2 & 1 & 78.6  & 215.3 & 0.20 & 36 \\
\bottomrule 
\end{tabular}

%% file: tables/relocalization.tex
\setlength{\tabcolsep}{3pt}
\renewcommand{\arraystretch}{0.9}
\begin{tabular}{L{1cm} C{0.7cm} C{0.8cm} C{1.2cm} L{0.7cm} C{0.7cm} C{1.2cm} C{0.6cm}}
\toprule
\textbf{Mission} & \multicolumn{3}{l}{\textbf{Relocalizations}} & \multicolumn{4}{l}{\textbf{Distance Between Reloc.}~[$m$]}\\
\textbf{Name} & \textbf{Scans} & \textbf{Succ.} & \textbf{Rate}~[$\%$] & \textbf{Path} & \textbf{Mean} & \textbf{Median} & \textbf{Max}\\
\midrule
\textsf{Seq-01} & 269  & 59  & 21.9  & 119.8  & 2.0  & 0.7 & 25.6 \\
\textsf{Seq-02} & 512  & 68  & 13.3  & 98.9   & 1.4  & 0.8 & 28.4 \\
\textsf{Seq-03} & 456  & 162 & 35.5  & 175.4  & 1.0  & 0.6 & 18.3 \\
\bottomrule 
\end{tabular}